\documentclass[10pt,journal,compsoc]{IEEEtran}
\usepackage{amsmath,amsfonts}
\usepackage{algorithm}
\usepackage{algorithmicx}
\usepackage{algpseudocode}
\usepackage[caption=false,font=normalsize,labelfont=sf,textfont=sf]{subfig}
\usepackage{textcomp}
\usepackage{stfloats}
\usepackage{url}
\usepackage{verbatim}
\usepackage{graphicx}
\usepackage{times}
\usepackage{epsfig}

\usepackage{amssymb}
\usepackage{dsfont}
\usepackage{array}
\usepackage{booktabs}
\usepackage{pifont}
\usepackage{multirow}
\usepackage{isomath,commath,multirow}
\usepackage{tikz}
\usepackage{color, colortbl,xcolor}
\definecolor{Gray}{gray}{0.9}
\def\*#1{\mathbf{#1}}
\def\^#1{\mathrm{#1}}

\usepackage{tikz}

\ifCLASSOPTIONcompsoc
  \usepackage[nocompress]{cite}
\else
  \usepackage{cite}
\fi

\ifCLASSINFOpdf
\else
\fi

\begin{document}

\title{Reversible Graph Neural Network-based Reaction Distribution Learning for Multiple Appropriate Facial Reactions Generation}

\author{Tong Xu,
        Micol Spitale,
        Hao Tang,
        Lu Liu,
        Hatice Gunes, 
        and
        Siyang Song$^*$

\IEEEcompsocitemizethanks{\IEEEcompsocthanksitem Tong Xu, Lu Liu and Siyang Song are with the School of Computing and Mathematical Sciences, University of Leicester, Leicester, LE2 7RH, United Kingdom.
E-mail:  l.liu@leicester.ac.uk, ss1535@leicester.ac.uk \\
($^*$ Corresponding Author: Siyang Song, E-mail: ss1535@leicester.ac.uk) \\
\indent Micol Spitale, Hatice Gunes and Siyang Song are with the AFAR Lab, Department of Computer Science and Technology, University of Cambridge, Cambridge, CB3 0FT, United Kingdom.
E-mail:  ms2871@cam.ac.uk, Hatice.Gunes@cl.cam.ac.uk, ss2796@cam.ac.uk \\
\indent Hao Tang is with the Department of Information Technology and Electrical Engineering, ETH Zurich,  Zurich 8092, Switzerland. E-mail: hao.tang@vision.ee.ethz.ch}

\thanks{Manuscript received June 15, 2023}}

\markboth{IEEE TRANSACTIONS on AFFECTIVE COMPUTING}%
{Shell \MakeLowercase{\textit{et al.}}: Bare Advanced Demo of IEEEtran.cls for IEEE Computer Society Journals}

\IEEEtitleabstractindextext{%
\begin{abstract}
Generating facial reactions in a human-human dyadic interaction is complex and highly dependent on the context since more than one facial reactions can be appropriate for the speaker's behaviour.
This has challenged existing machine learning (ML) methods, whose training strategies enforce models to reproduce a specific (not multiple) facial reaction from each input speaker behaviour.
This paper proposes the first multiple appropriate facial reaction generation framework that re-formulates the \textit{one-to-many mapping} facial reaction generation problem as a \textit{one-to-one mapping} problem. This means that we approach this problem by considering generating a distribution of listeners' appropriate facial reactions instead of multiple different appropriate facial reactions,, i.e., 'many' appropriate facial reaction labels are summarised as 'one' distribution label during training. Our model consists of a perceptual processor, a cognitive processor, and a motor processor. The motor processor is implemented with a novel Reversible Multi-dimensional Edge Graph Neural Network (REGNN). This allows us to obtain a distribution of appropriate real facial reactions during the training process, enabling the cognitive processor to be trained to predict the appropriate facial reaction distribution. At the inference stage, the REGNN decodes an appropriate facial reaction by using the predicted distribution as input.
Experimental results demonstrate that our approach outperforms existing models in generating more appropriate, realistic, and synchronized facial reactions. The improved performance is largely attributed to the proposed appropriate facial reaction distribution learning strategy and the use of a REGNN. The code will be made publicly available at \url{https://github.com/TongXu-05/REGNN-Multiple-Appropriate-Facial-Reaction-Generation}.


\end{abstract}

}

\maketitle

\IEEEdisplaynontitleabstractindextext

%
\IEEEpeerreviewmaketitle

\ifCLASSOPTIONcompsoc
\IEEEraisesectionheading{\section{Introduction}\label{sec:introduction}}
\else
\section{Introduction}
\label{sec:introduction}
\fi

\noindent \IEEEPARstart{N}{on-verbal} behaviour interaction plays a key role in human-human communication, with facial reactions providing important cues for understanding each other's intentions as well as affective and emotional states \cite{MEANING}. In dyadic interactions, a facial reaction refers to the \textbf{listener}'s non-verbal facial behaviours in response to the \textbf{speaker}'s verbal and non-verbal behaviours (e.g., facial muscle movements) \cite{theory1,mehrabian1974approach}. Previous studies \cite{bio1,bio2} have shown that the generation of listener's facial reactions to a speaker's behaviour in dyadic interaction consists of three main stages: Firstly, the listener's perceptual system (e.g., ears and eyes) receives external signals expressed by the speaker, which are pre-processed before being transmitted to the brain for further analysis. Then, the cognitive processor processes the pre-processed signals by taking personalized perception bias into account, resulting in the generation of personalized reaction signals, which is also influenced by various internal disposes (e.g., emotional states \cite{manstead1999social}). Finally, the motor processor decodes these personalized signals to the facial muscles, producing corresponding facial reactions. 

\begin{figure}[tp]
	\begin{center}
		\includegraphics[width=1.0\hsize]{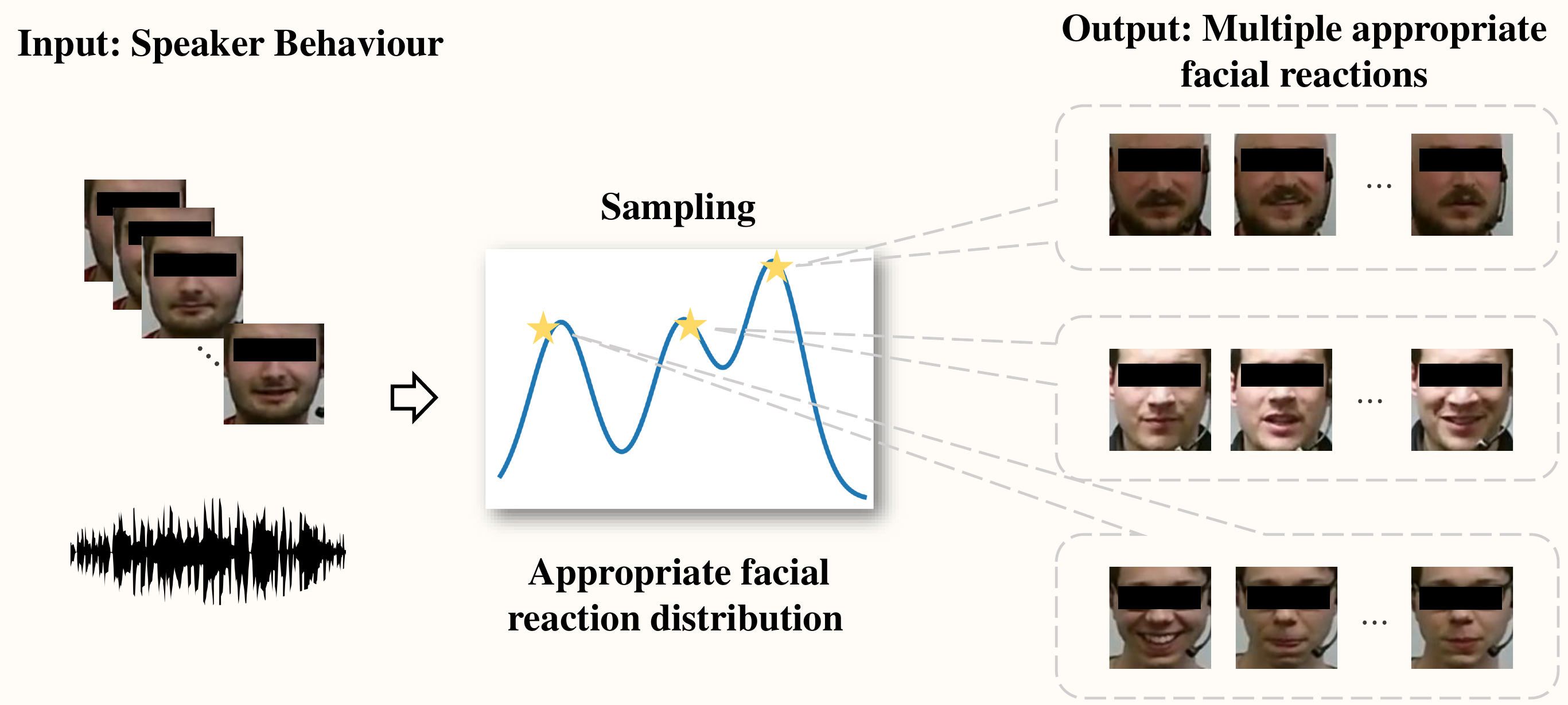}
	\end{center}
	\caption{Our approach predicts an distribution representing multiple different but appropriate facial reactions from each input speaker behaviour, based on which multiple different but appropriate, realistic, and synchronized human listener facial reactions could be generated.}
	\label{fig:show}
\end{figure}

In contrast to most 'one-to-one mapping' facial machine learning tasks (e.g., face recognition), the generation of listener's facial reactions to a specific speaker behaviour are characterized by variability and uncertainty \cite{song2023multiple,ng2022learning}. Specifically, given a speaker behaviour, different facial reactions can be expressed across not only different human listeners but also the same individual under different conditions (e.g., different emotional states and external environments), i.e., \textit{multiple different facial reactions could be appropriate in response to a speaker behaviour}. Most existing machine learning (ML)-based Facial Reaction Generation (FRG) models aim to reproduce the real facial reaction expressed by the corresponding listener under a specific context (called ``GT reaction'' in this paper) in response to the given speaker behaviour. These models -- including Generative Adversarial Networks (GAN) \cite{CGAN-BMVC,CGAN-CVPR}, VQ-VAE \cite{ng2022learning}, and person-specific FRG networks \cite{song2022learning,shao2021personality} -- are trained by minimizing L1 or L2 loss between the generated and GT real facial reactions. However, this 'one-to-one mapping' training strategy creates an ill-posed problem for existing FRG models where similar inputs (speaker behaviours) are paired with different labels (listener facial reactions), resulting in a ``one-to-many mapping'' problem in the training phase. This limitation makes it theoretically very challenging for the aforementioned approaches to learn good FRG models for generating appropriate and realistic facial reactions in response to speaker behaviours exhibited within various listener and contextual settings.


In this paper, we propose the first deep learning framework for generating multiple appropriate, diverse, and realistic facial reactions in response to each speaker behaviour, which specifically address the ``one-to-many mapping'' problem during the training phase. Inspired by the theoretical framework of the human model processor \cite{card1986model}, our approach is designed to consist of three modules: (i) a \textbf{perceptual processor} that encodes a pair of representations describing input speaker audio and facial signals; (ii) a \textbf{cognitive processor} that predicts an appropriate facial reaction distribution from the encoded speaker audio and facial representations, representing multiple different but appropriate facial reactions in response to the input speaker behaviour; and (iii) a reversible Graph Neural Network (GNN)-based \textbf{motor processor} that decodes an appropriate facial reaction from the learned distribution.

To address the ``one-to-many mapping'' problem, we propose a novel Reversible Multi-dimensional Edge Graph Neural Network (REGNN) as the motor processor. In the training phase, the REGNN summarises a distribution (called appropriate real facial reaction distribution in this paper) for each speaker behaviour. This distribution represents all real facial reactions in the training set, which are considered to be appropriate in response to this speaker behaviour. Then, the obtained appropriate real facial reaction distribution is employed to supervise the training process of the cognitive processor, by enforcing it to output the same distribution from the input speaker behaviour. As a result, this distribution learning strategy reformulates the ill-posed ``one-to-many mapping'' training problem, where one input speaker behaviour corresponds to multiple appropriate listener facial reactions, into a well-posed ``one-to-one mapping'' training problem, where one input speaker behaviour corresponds to one appropriate facial reaction distribution. Moreover, our graph-based model allows the relationship between each pair of facial attribute in the predicted facial reaction to be explicitly modelled via the corresponding multi-dimensional edge features, which further enhanced the quality of the generated facial reactions. We illustrate our approach in Fig. \ref{fig:show} and Fig. \ref{fig:Framework}. The main contributions and novelties of this paper are summarised as follows:
\begin{itemize}
    \item To the best of our knowledge, we present the first deep learning framework capable of generating multiple appropriate, realistic, and synchronized facial reactions in response to a speaker behaviour. Our framework introduces a novel appropriate facial reaction distribution learning (AFRDL) strategy that reformulates the ill-posed ``one-to-many mapping'' training problem occurring in existing facial reaction generation approaches as a ``one-to-one mapping'' problem, thus providing a well-defined learning objective.

    \item We propose a novel Reversible Multi-dimensional Edge Graph Neural Network (REGNN) for our facial reaction distribution learning. The REGNN can forwardly summarise a distribution from multiple real appropriate facial reactions at the training stage, and reversely decode multiple appropriate facial reactions from the predicted facial reaction distribution at the inference stage.

    \item We generated -- using the proposed approach --  appropriate, realistic, and synchronized facial reactions achieving better performances compared to other existing related solutions, and we provide the first open-source code for the multiple appropriate facial reaction generation task.

\end{itemize}

\section{Related Work}

\noindent In this section, we first review biological and psychological studies that explain human facial reaction mechanisms in Sec. \ref{subsec:Facial Reaction Theory}. Then, existing ML-based facial reaction generation approaches are summarised in Sec. \ref{subsec:Automatic facial reaction}. 


\subsection{Facial reaction theory}
\label{subsec:Facial Reaction Theory}




\noindent During dyadic interactions, the facial reactions of a listener are shaped by a combination of facial muscle movements. These movements are controlled by person-specific cognitive processes that are primarily influenced by the behaviours expressed by the corresponding speaker \cite{dimberg1990distinguished}. Research conducted by Hess et al. \cite{hess1998facial} also found that the generation of facial reactions is predominantly influenced by individual-specific cognitive processes, which are not only influenced by the speaker's behaviour but also by the listener's personality \cite{lang1993looking} and emotional states \cite{dimberg1982facial}. For instance, individuals who frequently experience fear possess more sensitive and easily stimulated amygdalae, rendering them more prone to displaying facial reactions indicative of fear. Similarly, experiencing pleasant emotions triggers the contraction of the zygomatic major muscle, resulting in a smiling facial reaction, while confusion enhances the activity of the corrugator muscle, leading to a furrowed brow expression. Therefore, as summarised in \cite{song2023multiple}, in dyadic interactions, a broad spectrum of different facial reactions might be \textit{appropriate} in response to a speaker behaviour according to the internal states of the listener. This is because human behavioural responses are stimulated by the context the listener experiences \cite{mehrabian1974approach}, which lead to different but appropriate facial reactions expressed by not only different listeners but also the same listener under different contexts (e.g., external environments or internal states) \cite{song2023multiple,zhai2020sor,pandita2021psychological}. A similar hypothesis has been mentioned in a recent facial reaction generation study \cite{ng2022learning}.

\subsection{Automatic facial reaction generation}
\label{subsec:Automatic facial reaction}

\noindent To the best of our knowledge, there are only a few studies \cite{CGAN-BMVC, CGAN-CVPR,CGAN-ICMI,song2022learning,shao2021personality,nojavanasghari2018interactive,ng2022learning,listeninghead} have been investigated automatic facial reaction generation task. An early approach \cite{CGAN-CVPR} proposed a two-stage conditional GAN to generate facial reaction sketches based on the speaker's facial action units (AUs). Their later works \cite{CGAN-BMVC,CGAN-ICMI} exploited more speaker emotion-related features (e.g., facial expression features) to reproduce better facial reactions expressed by listeners. Similar strategy \cite{woo2021creating,woo2023amii,L2L,geng2023affective} have been extended for the same purpose, where all these approaches directly measure the similarity between each generated facial reaction with the specific facial behaviour expressed by the corresponding human listener (i.e., the model is trained to reproduce corresponding real facial reactions based on speaker behaviours expressed by different subjects under various conditions). For example, Geng et al. \cite{geng2023affective} leveraged pre-trained large language models (LLM) and vision-language models to generate the best reaction to the speaker's speech behaviour. To consider personalized factors in expressing facial behaviours, Song et al. \cite{song2022learning,shao2021personality} used Neural Architecture Search (NAS) to explore a person-specific network for each listener adaptively, and thus each network is specifically explored to reproduce the corresponding listener's facial reactions. Despite of this 'one-to-one mapping' training strategy, Ng et al. \cite{L2L} extended and combined the cross-attention transformer with VQ-variational auto-encoder (VQ-VAE) \cite{VQ-VAE} model, allow a range of diverse facial reactions can be generated from each input multi-modal speaker behaviour. Based on the interlocutor's speech and facial motion, the approach proposed by Jone et al. \cite{jonell2020let} can sample multiple avatar's facial reactions depsite their appropriateness is not objectively measured. To the best of our knowledge, none of existing publications have attempted to generate multiple \textbf{appropriate} facial reactions from each speaker behaviour (please refer to \cite{song2023multiple} for detailed task definition).

Note that the approach proposed in this paper is different from previous facial expression/display generation methods \cite{ganimation,otberdout2020dynamic,otberdout2022sparse,fan2022faceformer,richard2021meshtalk,tang2023bipartite,tang2019expression,tang2022facial}, where the facial images are generated based on manually defined conditions such pre-defined AUs, landmarks and audio behaviours without considering interaction scenarios (i.e., they do not predict reactions from speaker behaviours).










\section{Task definition}
\label{sec:task definition}

\noindent Given a speaker behaviour $B_S^{t_1, t_2} = \{ A_{S}^{t_1,t_2}, F_{S}^{t_1,t_2} \}$ at the time $[t_1, t_2]$, the goal is to learn a ML model $\mathcal{H}$ that can generate multiple different spatio-temporal human facial reactions $P_L(B_S^{t_1, t_2}) = \{ p_L(B_S^{t_1, t_2})_1, \cdots, p_L(B_S^{t_1, t_2})_N  \}$ that are \textbf{appropriate} in response to $B_S^{t_1, t_2}$, which can be formulated as:
\begin{equation}
   P_L(B_S^{t_1, t_2}) = \mathcal{H}(B_S^{t_1, t_2}),
\label{eq:problem}
\end{equation}
where $p_L(B_S^{t_1, t_2})_1 \neq \cdots \neq p_L(B_S^{t_1, t_2})_N$. Here, each generated facial reaction $p_L(B_S^{t_1, t_2})_n$ ($n = 1, 2, \cdots, N$) should be similar to at least one real facial reaction $f_L(B_S^{t_1, t_2})_m$ that is appropriate in response to $B_S^{t_1, t_2}$ in the training set:
\begin{equation}
   p(F_L \vert B_S^{t_1, t_2})_n \approx f_L(B_S^{t_1, t_2})_m \in  F_L(B_S^{t_1, t_2}),
\end{equation}
where $F_L(B_S^{t_1, t_2}) = \{ f_L(B_S^{t_1, t_2})_1, \cdots, f_L(B_S^{t_1, t_2})_M \}$ denotes a set of real facial reactions expressed by human listeners in the training set, which are appropriate in response to the speaker behaviour $B_S^{t_1, t_2}$. The above definition corresponds to the offline multiple appropriate facial reaction generation task (offline MAFRG) defined by \cite{song2023multiple} and an open challenge \footnote{https://sites.google.com/cam.ac.uk/react2023/home}.

\begin{figure*}[tp]
	\begin{center}
		\includegraphics[width=0.96\textwidth]{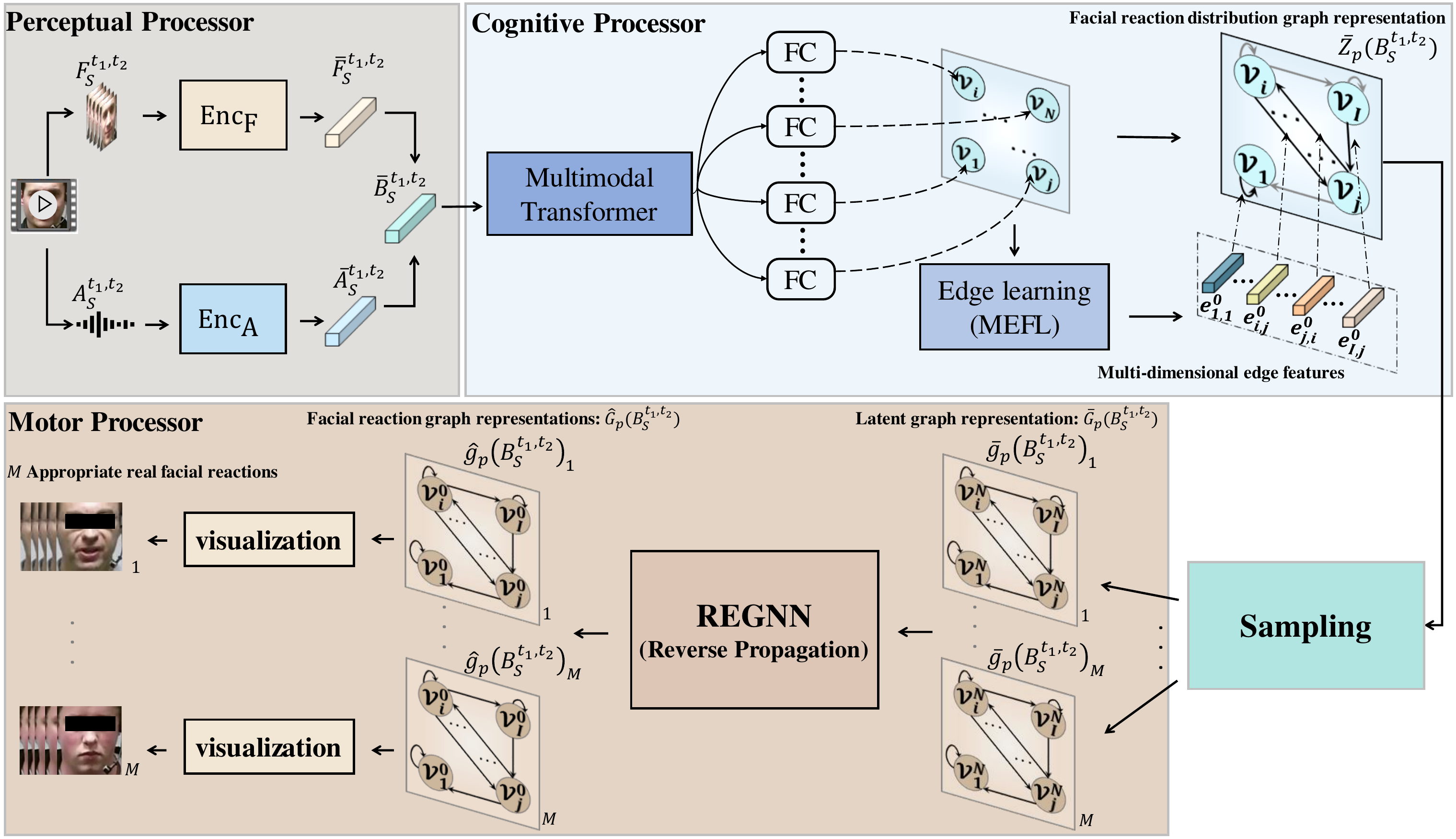}
	\end{center}
	\caption{Overview of the proposed multiple appropriate facial reaction generation framework. \textbf{Step 1:} the \textbf{Perceptual Processor} first encodes facial and audio representations from the perceived audio-visual speaker behaviours. \textbf{Step 2:} the \textbf{Cognitive Processor} then predicts a distribution from the combined audio-visual representation, which represents all appropriate facial reactions in response to the input speaker behaviour. \textbf{Step 3:} the REGNN-based \textbf{Motor Processor} finally samples and reversely decodes (\textbf{reverse propagation}) multiple appropriate facial reactions from the learned distribution.
	}
	\label{fig:Framework}
\end{figure*}

\section{The proposed approach}
\label{sec:approach}

\noindent This section presents the details of our MAFRG approach. The whole pipeline is firstly introduced in Sec. \ref{subsec:pipeline}, and then the proposed appropriate facial reaction distribution learning (AFRDL) strategy is explained in Sec. \ref{subsec:training}. Finally, we provide the details of our REGNN in Sec. \ref{subsec:Reversible GNN}, which plays a key role in the AFRDL strategy.

\subsection{Facial reaction generation framework}
\label{subsec:pipeline}

\noindent The proposed MAFRG model $\mathcal{H} = \{ \textbf{Enc}, \textbf{Cog}, \textbf{Mot} \}$ aims to generate multiple diverse and appropriate human facial reactions $P_L(B_S^{t_1, t_2}) = \{ p_L(B_S^{t_1, t_2})_1, \cdots, p_L(B_S^{t_1, t_2})_N  \}$ in response to each speaker audio-facial behaviour $B_S^{t_1,t_2} = \{  A_S^{t_1,t_2},  F_S^{t_1,t_2}\}$. As shown in Fig. \ref{fig:Framework}, our model consists of three main modules inspired by the Human Model Processor (HMP) \cite{card1986model}: (i) \textbf{Perceptual Processor} ($\textbf{Enc} = \{ \textbf{Enc}_\text{A}, \textbf{Enc}_\text{F} \}$) that encodes each raw speaker audio-facial behaviour into a pair of latent audio and facial representations; (ii) \textbf{Cognitive Processor ($\text{Cog}$}) that predicts a distribution representing all appropriate facial reactions in response to $B_S^{t_1,t_2}$, based on the produced speaker audio and facial representations; and (iii) the REGNN-based \textbf{Motor Processor ($\textbf{Mot}$)} that samples and generates an appropriate facial reaction from the predicted distribution. The pipeline of our model is illustrated in Fig. \ref{fig:Framework}.

\textbf{Perceptual Processor.} The \textbf{Perceptual Processor} is a two branch encoder consisting of a facial encoder $\textbf{Enc}_\text{F}$ (Swin-Transformer \cite{swin}) and an audio encoder $\textbf{Enc}_\text{A}$ (VGGish \cite{hershey2017cnn}). It takes the speaker audio and facial signals $A_S^{t_1,t_2}$ and $F_S^{t_1,t_2}$ expressed at the time interval $[t_1,t_2]$ as the input, and generates a pair of latent audio and facial representations $\bar A_S^{t_1,t_2}$ and $\bar F_S^{t_1,t_2}$ as:
\begin{equation}
\begin{split}
     & \bar A_S^{t_1,t_2} = \textbf{Enc}_\text{A}(A_S^{t_1,t_2}) \\
     & \bar F_S^{t_1,t_2} = \textbf{Enc}_\text{F}(F_S^{t_1,t_2})       
\end{split}
\end{equation}

\textbf{Cognitive Processor.} Based on the learned representations $\bar A_S^{t_1,t_2}$ and $\bar F_S^{t_1,t_2}$, the \textbf{Cognitive Processor} first aligns and combines them as a latent speaker audio-facial behaviour representation $\bar{B}_S^{t_1,t_2}$ using the same attention-based strategy introduced in \cite{multimodal}. Specifically, instead of predicting a specific facial reaction from the speaker behaviour, we propose to predict an \textit{appropriate facial reaction distribution graph representation} $\bar{Z}_p(B_S^{t_1,t_2})$ representing multiple facial reactions that are appropriate for responding to $\bar{B}_S^{t_1,t_2}$. This process is achieved through the use of $I$ projection heads (i.e., $I$ fully connected (FC) layers), where each head learns a $D$-dimensional vector that is specifically treated as a node feature for $\bar{Z}_p(B_S^{t_1,t_2})$, which can be formulated as:
\begin{equation}
    \bar{Z}_p(B_S^{t_1,t_2}) = \textbf{COG}(\bar{B}_S^{t_1,t_2}),
\label{eq:cog-processor}
\end{equation}
where $\bar{Z}_p(B_S^{t_1,t_2}) \in \mathbb{R}^{I \times D}$ ($I$ nodes). Here, each node represents the distribution of a specific facial attribute time-series of the predicted reaction (please see Sec. \ref{subsec:training} for details). This way, \textit{the ``one-to-many mapping'' problem occurring at the training phase in the FRG task is addressed by re-formulating it into a ``one-to-one mapping'' problem (one speaker behaviour corresponds to one appropriate facial reaction distribution).} Then, we feed all node features to our multi-dimensional edge feature learning (MEFL) block that consists of $D$ attention operations, where each attention operation generates an $I \times I$ attention map describing a specific type of mutual relationship between a pair of nodes. Consequently, $D$ attention maps describing $D$ types of relationship cues would be produced. Thus, for each pair of nodes, a pair of multi-dimensional directed edge features can be obtained to describe their relationship, i.e., each multi-dimensional edge feature $e^{\bar{Z}_p}_{i,j}$ is a $D$ dimensional vector produced by concatenating the values at the $i_\text{th}$ row and $j_\text{th}$ column of $D$ attention maps. Here, we only keep $\mathcal{K}$ directed edges starting from each node, which has top-K largest norms.

\textbf{Motor Processor.} To allow facial attributes of an appropriate facial reaction to be generated from the distribution $\bar{Z}_p(B_S^{t_1,t_2})$, we propose a novel Reversible Multi-dimensional Edge Graph Neural Network (REGNN) as the \textbf{Motor Processor}. In particular, as human facial behaviours can be interpreted by various medium and high level primitives (e.g., facial Action Units (AUs) and affects) that are mutual correlated, we propose to describe each spatio-temporal facial reaction as $I$ facial attribute time-series which are further represented as a graph consisting of $I$ nodes. The presence of each directed edge connecting nodes and the corresponding multi-dimensional edge feature are also defined by the MEFL block. Consequently, the relationship between each pair of facial attribute time-series (nodes) in the predicted facial reaction can be explicitly described by a pair of multi-dimensional edge features during the reasoning. To generate aa appropriate facial reaction, the proposed REGNN first samples an appropriate facial reaction latent graph representation $\bar{g}_p(B_S^{t_1,t_2})_n \in \mathbb{R}^{I \times D}$ from the predicted distribution $\bar{Z}_p(B_S^{t_1,t_2})$, and then decodes it as a facial reaction graph representation $\hat{g}_p(B_S^{t_1,t_2})_n \in \mathbb{R}^{I \times D}$. Here, multiple different facial reaction latent graph representations can be sampled from $\bar{Z}_p(B_S^{t_1,t_2})$, and thus the multiple corresponding multiple different facial reactions $\hat{G}_p(B_S^{t_1,t_2}) = \{\hat{g}_p(B_S^{t_1,t_2})_1, \cdots, \hat{g}_p(B_S^{t_1,t_2})_N\}$ can be also generated. This can be formulated as:
\begin{equation}
     \hat{g}_p(B_S^{t_1,t_2})_n  = \textbf{Mot}^{-1}(Z_p(B_S^{t_1,t_2})),
\label{eq: motor-processor}
\end{equation}
where $\text{Mot}^{-1}$ denotes that the motor processor reversely infers a facial reaction from the predicted distribution $Z_p(B_S^{t_1,t_2})$. Subsequently, a 2D facial reaction image sequence $p_L(B_S^{t_1, t_2})_n$ can be further produced from $\hat{g}_p(B_S^{t_1,t_2})_n$.

\subsection{Appropriate facial reaction distribution learning}
\label{subsec:training}

\begin{figure*}[tp]
	\begin{center}
		\includegraphics[width=1.0\hsize]{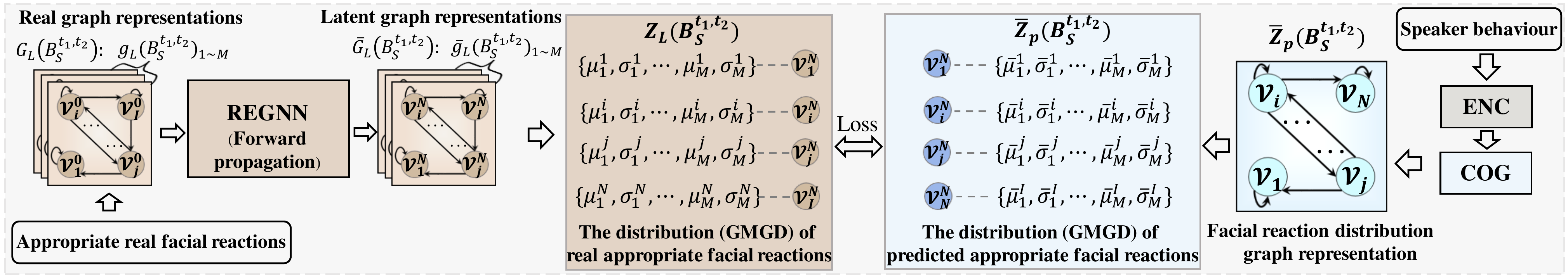}
	\end{center}
	\caption{Illustration of the proposed AFRDL strategy. Given a speaker behaviour, REGNN first encodes all appropriate real facial reactions as a set of latent graph representations. These representations are then summarised as an appropriate real facial reaction distribution \textbf{(forward propagation}) to supervise the Cognitive Processor's training, where the summarised distribution is a graph representation consisting of multiple nodes, and each node is represented by a Gaussian Mixture Model (GMM) summarising multiple facial attribute time-series corresponding to multiple appropriate real facial reactions. Here, the MSE loss function is employed to enforce the distribution predicted by the Cognitive Processor to be similar to the summarised appropriate real facial reaction distribution.}
	\label{fig:Distribution}
\end{figure*}




\noindent Our appropriate facial reaction distribution learning (AFRDL) strategy aims to address the ``one-to-many mapping'' problem occurring in FRG models' training (i.e., one input speaker behaviour corresponds to multiple appropriate facial reaction labels) by re-formulating it as a ``one-to-one mapping'' problem (i.e., one input speaker behaviour corresponds to one distribution representing multiple appropriate facial reactions). To enforce the \textbf{Cognitive Processor} accurately predicting the distribution of all appropriate facial reactions in response to each input speaker behaviour, we apply our REGNN to learn an appropriate real facial reaction distribution graph representation $Z_L(B_S^{t_1,t_2})$ for each speaker behaviour $B_S^{t_1,t_2}$, representing the distribution of all real facial reactions that are appropriate for responding to $B_S^{t_1,t_2}$. The obtained distributions are then treated as the targets to supervise the training process of the \textbf{Cognitive Processor}. Here, the appropriate real facial reactions for responding to each speaker behaviour are defined based on the automatic and objective labelling strategy provided in \cite{song2023multiple}.

As shown in Fig. \ref{fig:Distribution}, given an audio-visual speaker behaviour $B_S^{t_1,t_2} = \{A_S^{t_1,t_2}, F_S^{t_1,t_2} \}$, and its corresponding multiple appropriate real facial reactions $F_L(B_S^{t_1, t_2})$ expressed by human listeners in the training set, we first construct a set of real facial reaction graph representations $G_L(B_S^{t_1,t_2}) = \{g_L(B_S^{t_1,t_2})_1, \cdots, g_L(B_S^{t_1,t_2})_M \}$ that represent all appropriate real facial reactions defined by $F_L(B_S^{t_1, t_2})$, where each node in a graph representation describes a facial attribute time-series and each multi-dimensional edge feature defined by the MEFL explicitly describes the relationship between a pair of nodes. Since all graphs in $G_L(B_S^{t_1,t_2})$ have a common property, i.e., all of them describe appropriate facial reactions in response to $B_S^{t_1,t_2}$, we hypothesize that they are drawn from the same distribution. Subsequently, we train the REGNN by enforcing it to map all appropriate real facial reaction graph representations in response to the same speaker behaviour onto a ``ground-truth'' (GT) real appropriate facial reaction distribution $Z_L(B_S^{t_1,t_2})$ as:
\begin{equation}
\begin{split}
    &\bar{g}_L(B_S^{t_1,t_2})_m  = \textbf{Mot}(g_L(B_S^{t_1,t_2})_m), \\ 
    &\bar{g}_L(B_S^{t_1,t_2})_m \sim Z_L(B_S^{t_1,t_2}), \quad m = 1, 2, \cdots M \\
    & \textbf{subject to} ~~ f_L(B_S^{t_1, t_2})_m \in F_L(B_S^{t_1, t_2})
\end{split}
\end{equation}
where $\bar{g}_L(B_S^{t_1,t_2})_m$ denotes a latent graph representation produced from $g_L(B_S^{t_1,t_2})_m$, and all latent graph representations are expected to follow the same distribution $Z_L(B_S^{t_1,t_2})$. The training process is achieved by minimizing the sum of L1 distances obtained from all the corresponding latent graph representation pairs in an unsupervised manner:
\begin{equation}
    \mathcal{L}_{1} = \sum_{m_1=1}^{M-1} \sum_{m_2=m_1+1}^{M} L1(\hat{g}_L(B_S^{t_1,t_2})_{m_2}, \hat{g}_L(B_S^{t_1,t_2})_{m_1})
\end{equation}

Inspired by the fact that Guassian Mixture Model (GMM) is powerful to describe distributed subpopulations (e.g., individual appropriate facial reaction in our case) within an overall population (e.g., all appropriate facial reactions), we propose a novel \textbf{Gaussian Mixture Graph Distribution (GMGD)} to represent $Z_L(B_S^{t_1,t_2}) = \{v_1^Z, v_2^Z, \cdots, v_I^Z\}$, where each node $v_i^Z$ is represented by a Gaussian Mixture Model (GMM) consisting  of $M = D/2$ Gaussian distributions (defined as $\mathcal{N}(\{\mu^1_i, \cdots, \mu^M_i\}, \{\sigma^1_i, \cdots, \sigma^M_i\})$). Specifically, for the $i_\text{th}$ node $v_i^Z \in Z_L(B_S^{t_1,t_2})$, the $M$ mean values $\mu^1_i, \cdots, \mu^M_i$ corresponding to its $M$ Gaussian distributions are defined by the $M$ latent graph representations $\bar{g}_L(B_S^{t_1,t_2})_1, \cdots, \bar{g}_L(B_S^{t_1,t_2})_M$ produced by the motor processor, i.e., the $i_\text{th}$ node features of these $M$ latent graph representations, which can be formulated as:
\begin{equation}
    \mu^m_i = \bar{v}(L)_i^m \in \bar{g}_L(B_S^{t_1,t_2})_m, ~~ m = 1,2, \cdots, M
\end{equation}
where $\bar{v}(L)_i^m$ is the $i_\text{th}$ node feature in the $m_\text{th}$ latent graph representation $\bar{g}_L(B_S^{t_1,t_2})_m$. Meanwhile, standard deviations $\{\sigma(L)^1_i, \cdots, \sigma(L)^M_i\})$ are empirically defined ($\sigma(L)^1_i = \cdots = \sigma(L)^M_i = 0.06$ is used in this paper).

As a result, the \textbf{Cognitive Processor} is trained under the supervision of the appropriate real facial reaction distribution $Z_L(B_S^{t_1,t_2})$ produced by the REGNN, aiming to predict an appropriate facial reaction distribution graph representation $\bar{Z}_p(B_S^{t_1,t_2}) = Z_L(B_S^{t_1,t_2})$ from $B_S^{t_1,t_2}$ (formulated in Eq. \eqref{eq:cog-processor}). This training process is achieved by minimizing the L2 distance between $Z_L(B_S^{t_1,t_2})$ and $\bar{Z}_L(B_S^{t_1,t_2})$ as:
\begin{equation}
\begin{split}
    \mathcal{L}_{2} = \text{MSE}(\bar{Z}_p(B_S^{t_1,t_2}), Z_L(B_S^{t_1,t_2}))
\end{split}
\end{equation}
where MSE denotes the Mean Square Error. At the inference stage, the well-trained REGNN first samples a facial reaction graph representation $\bar{g}_p(B_S^{t_1,t_2})_n$ from the predicted distribution $\bar{Z}_p(B_S^{t_1,t_2})$, and then reversely decodes it as a facial reaction graph representation.

\subsection{Reversible Multi-dimensional Edge Graph Neural Network (REGNN)}
\label{subsec:Reversible GNN}

\begin{algorithm*}
    \caption{Forward Propagation}
    \label{algorithm: 1}
    \begin{algorithmic}[1]
        \Ensure{the node feature set $\mathcal{V}^{n-1}$, \ \ the initial edge feature set $\mathcal{E}^{0}$}
        \Require{$\mathcal{V}^{n}$}
        
        \State{${\bar{\mathcal{V}}^{n-1}} = \mathrm{norm}(\mathcal{V}^{n-1})$}
        \Comment{\textcolor{gray}{Performing normalization for all node features}}
        
        \State{${\hat{\mathcal{V}}^{n-1}} = \mathrm{Sig}({\bar{\mathcal{V}}^{n-1}})$}
        \Comment{\textcolor{gray}{Performing Sigmoid activation for all node features}}
        
        \State{$\mathcal{E}^{n} = \mathrm{Edge\ Update}(\hat{\mathcal{V}}^{n-1}, \mathcal{E}^{0})$}
        \Comment{\textcolor{gray}{Updating all edge features based on the Eq. \ref{eq: edge updating}}}
        
        \State{${\mathcal{V}^{n-1}}' = \mathrm{Message\ Passing}
        (\hat{\mathcal{V}}^{n-1}, \mathcal{E}^{n})$}
        \Comment{\textcolor{gray}{Aggregating messages from adjacent nodes for every node based on the Eq. \ref{eq: message passing}}}
        
        \State{$\mathcal{V}^{n} = \mathcal{V}^{n-1} + {\mathcal{V}^{n-1}}'$}
        \Comment{\textcolor{gray}{Updating all node features based on the Eq. \ref{eq:node updating}}}
    \end{algorithmic}
\end{algorithm*}

\begin{algorithm*}
    \caption{Reverse Propagation}
    \label{algorithm: 2}
    \begin{algorithmic}[1]
        \Ensure{the node feature set $\mathcal{V}^{n}$,\ \ the initial edge feature set $\mathcal{E}^{0}$}, the function $\varphi = \{\text{Sig}, \text{Edge Update}, \text{Message Passing}   \}$
        \Require{$\mathcal{V}^{n-1}$}
        \State{$k\gets0, \ x_k \gets rand()$}
        \Comment{\textcolor{gray}{Randomly initializing the node feature set $X_0$ for the first iteration}}
        
        \While {$k = 0$ or $X_k \neq X_{k-1}$}
        \Comment{\textcolor{gray}{When $X_k=X_{k-1}$, stopping the iteration}}
        
        \State {$\bar{X}_k = \mathrm{Sig}(X_k)$}
        \Comment{\textcolor{gray}{Performing Sigmoid activation for $X_k$}}
        
        \State{$\mathcal{E}^{n}_k = \mathrm{Edge\ Update}(X_k, \mathcal{E}^{0})$}
        \Comment{\textcolor{gray}{Updating all edge features based on the Eq. \ref{eq: edge updating}}}
        
        \State{$\hat{X}_k = \mathrm{Message\ Passing}
        (\bar{X}_k, \mathcal{E}^{n})$}
        \Comment{\textcolor{gray}{Aggregating messages from adjacent nodes for every node based on the Eq. \ref{eq: message passing}}}
        
        \State {$X_{k+1} = V_i^n - \hat{X}_k$}
        \Comment{\textcolor{gray}{Computing $X_{k+1}$}}
        
        \State {$k=k+1$}
        \EndWhile
        \State {$\mathcal{V}^{n-1} = \mathrm{Inv\ Norm}(X_k)$}
        \Comment{\textcolor{gray}{Inverse normalization}}
        
        
    \end{algorithmic}
\end{algorithm*}

\begin{figure}[tp]
	\begin{center}
		\includegraphics[width=1.0\hsize]{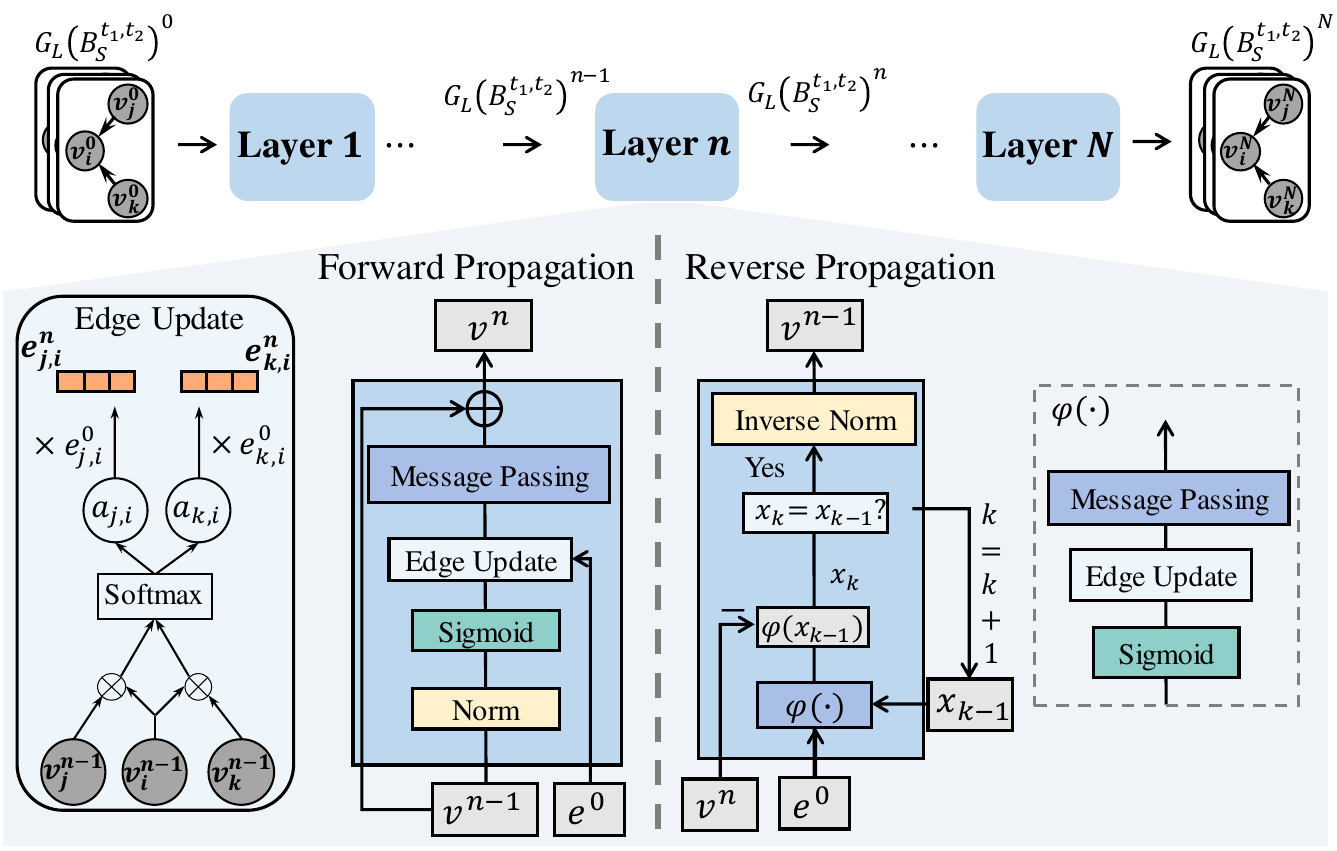}
	\end{center}
	\caption{The proposed Reversible Multi-dimensional Edge Graph Neural Network (REGNN) is made up of $N$ REGNN layers, where each layer can forwardly and reversely propagate node representations.}
	\label{fig:REGNN}
\end{figure}

\noindent In this paper, the proposed REGNN aims to \textbf{forwardly} encode a distribution that describes all appropriate real facial reactions in response to the input speaker behaviour, which plays a key role in supervising the cognitive processor's training. As shown in Fig. \ref{fig:REGNN}, the REGNN network consists of $N$ REGNN layers, which forwardly generates a graph $\mathcal{G}^N(\mathcal{V}^N,\mathcal{E}^N)$ from each input graph $\mathcal{G}^0(\mathcal{V}^0,\mathcal{E}^0)$, where $\mathcal{V}^0 = \{v_1^0, v_2^0, \cdots, v_I^0\}$ and $\mathcal{E} = \{e_{i,j}^0 \vert v_i^0, v_j^0 \in \mathcal{V} \quad \& \quad \mathcal A_{i,j} = 1  \}$ denote a set of node and edge features contained in the input graph $\mathcal{G}^0(\mathcal{V}^0,\mathcal{E}^0)$, respectively, while $\mathcal A$ is the adjacency matrix that defines the connectivity between nodes. Here, $\mathcal A$ remains the same for all latent and output graphs during the propagation. Importantly, our REGNN can also \textbf{reversely} output node feature set $\mathcal{V}^0$ from $\mathcal{V}^N$, as only nodes of the used graph representations contain target facial attributes/distributions.  The pseudo-code of the REGNN's the forward propagation and reverse propagation mechanisms are provided in Algorithm \ref{algorithm: 1} and Algorithm \ref{algorithm: 2}, respectively.

\textbf{Forward propagation:} The $n_\text{th}$ REGNN layer takes: (i) the node feature set $\mathcal{V}^{n-1}$ generated from the ${n-1}_{th}$ layer as its input node features; and (ii) the initial edge feature set $\mathcal{E}^0$ as its input edge features, and then outputs a graph $\mathcal{G}^n(\mathcal{V}^n,\mathcal{E}^n)$. As shown in Fig. \ref{fig:REGNN}, the node feature set $\mathcal{V}^n$ is learned by a normalization layer and a function $\varphi$ that consists of a Sigmoid activation, an edge updating operation and a message passing operation, which can be summarised as:
\begin{equation}
    \varphi = \{\text{Sig}, \text{Edge Update}, \text{Message Passing}   \}
\label{eq:phi}
\end{equation}
Subsequently, the forward propagation of the $n_\text{th}$ REGNN layer can be formulated as:
\begin{equation}
\begin{split}
    v_i^n = v_i^{n-1} + \varphi(\bar{v}_i^{n-1}) \\
    \bar{v}_i^{n-1} = \text{Norm}(v_i^{n-1})
\end{split}
\end{equation}
where $v_i^n \in \mathcal{V}^n$, $v_i^{n-1} \in \mathcal{V}^{n-1}$, and $\text{Norm}$ denotes the normalisation operation. Before formally updating edge and node features, the proposed REGNN also feeds normalised node features $\bar{v}_i^{n-1} \in \bar{\mathcal{V}}^{n-1}$ to an Sigmoid activation function, resulting in the activated node feature set $ \hat{\mathcal{V}}^{n-1} = \{ \hat{v}_1^{n-1}, \cdots, \hat{v}_I^{n-1} \}$, i.e., $\hat{v}_i^{n-1} = \text{Sig}(\bar{v}_i^{n-1})$. The use of the Sigmoid activation sets an upper-bound for the 2-norm value of the node feature vectors, which is a key step to ensure the REGNN's reversibility (please refer to Supplementary Material for details).

Subsequently, the $n_\text{th}$ REGNN layer learns edge feature set $\mathcal{E}^{n}$ based on not only the initial edge feature set $\mathcal{E}^0$ but also the activated node feature set $\hat{\mathcal{V}}^{n-1}$, allowing the obtained edge features to encode the latest relationships between nodes (i.e., $\hat{\mathcal{V}}^{n-1}$-related relationships). Specifically, the multi-dimensional feature $e^{n}_{j,i} \in \mathcal{E}^n$  of the directed edge starting from the node $\hat{v}_j^{n-1}$ to the node $\hat{v}_i^{n-1}$ is computed via the \textbf{Edge Update} operation as:
\begin{equation}
   e^{n}_{j,i} = \frac{a^{n}_{j, i} e^{0}_{j,i}}{\sum_{\hat{v}^{n-1}_k \in \mathcal N_{\hat{v}^{n-1}_i}} a^{n}_{k, i} e^{0}_{k, i}}
\label{eq: edge updating}
\end{equation}
where $e^{0}_{j,i} \in \mathcal{E}_{0}$ denotes an initial directed edge feature, and the term $\sum_{\hat{v}^{n-1}_k \in \mathcal N_{\hat{v}^{n-1}_i}} a^{n}_{k, i} e^{0}_{k, i}$ regularizes the obtained edge feature. Here, $a^{n}_{j,i}$ is a learnable relationship coefficient to define $\hat{\mathcal{V}}^{n-1}$-related and context-aware (i.e., aware of neighbouring nodes of $\hat{v}^{n-1}_i$) edge feature $e^{n}_{j,i}$. More specifically, $a^{n}_{j,i}$ is learned to capture relationships between the node $\hat{v}^{n-1}_i$ and its neighbouring nodes $\hat{v}^{n-1}_k \in \mathcal N_{\hat{v}^{n-1}_i}$, which can be computed as:
\begin{equation}
    a_{j,i}^{n} = \frac{\^{exp}\left((\hat{v}_{i}^{n-1}\*W_q^{n})(\hat{v}_{j}^{n-1}\*       
                        W_m^n)^{\top}\right)}
                       {\sum_{\hat{v}^{n-1}_k \in \mathcal N_{\hat{v}^{n-1}_i}} 
                       \^{exp}\left((\hat{v}_{i}^{n-1} \*W_q^{n})
                       (\hat{v}^{n-1}_k \* W_m^{n})^{\top}\right)}
\label{eq: coefficient}
\end{equation}
where $\* W_q^n$ and $\* W_m^n$ are learnable weight vectors of the $n_\text{th}$ REGNN layer, while $\mathcal N_{\hat{v}^{n-1}_i}$ denotes the adjacent node set of $\hat{v}_i^{n-1}$. Since $a_{j,i}^{n}$ and $e^{n}_{j,i}$ are learned to update the node feature $\hat{v}^{n-1}_i$ to $\hat{v}^{n}_i$, the proposed strategy encode $a_{j,i}^{n}$ to capture the information from not only the relationship cues between the corresponding node features $\hat{v}^{n-1}_j$ and $\hat{v}^{n-1}_i$ but also the context of the node $\hat{v}^{n-1}_i$ (its neighbouring nodes $\mathcal{N}_{\hat{v}^{n-1}_i}$).

Building upon the learned edge feature set $\mathcal{E}^{n}$, the $n_\text{th}$ REGNN layer then computes each node feature $v_i^{n}$ based on: (i) the node feature $\hat{v}_i^{n-1} \in \hat{\mathcal{V}}^{n-1}$; and (ii) the message $m_{\hat{v}_i^{n-1}}$ aggregated from all adjacent nodes $\hat{v}^{n-1}_j \in \mathcal{N}_{\hat{v}_i^{n-1}}$ of the $\hat{v}_i^{n-1}$ (i.e., these node features are also normalised and activated by Norm and Sig operations). This process can be formulated as:
\begin{equation}
    v_i^{n} = v_i^{n-1} + m_{\hat{v}_i^{n-1}}
\label{eq:node updating}
\end{equation}
Here, the message $m_{\hat{v_i^{n-1}}}$ is computed via the \textbf{Message Passing} operation, which is decided by not only the adjacent node feature set ($\mathcal{N}_{\hat{v}_i^{n-1}}$) but also the corresponding updated directed edges (i.e., multi-dimensional edge features $e^{n}_{j,i} \in \mathcal{E}^{n}$) pointing to the $\hat{v}_i^{n-1}$ as:  
\begin{equation}
      m_{\hat{v}_i^{n-1}} = \*W_e^{n} \sum_{\hat{v}^{n-1}_j \in \mathcal{N}_{\hat{v}_i^{n-1}}}\left(e_{j,i}^{n} \circ \hat{v}^{n-1}_j\right) 
\label{eq: message passing}
\end{equation}
where $\* W_e^{n} \in \mathbb{R}^{1 \times D}$ is a learnable weight vector that combines messages passed by all dimensions ($D$ dimensions) of the multi-dimensional edge $e_{j,i}^{n}$.

\textbf{Reverse propagation:} The $n_\text{th}$ REGNN layer can also reversely infer each input node feature $v_i^{n-1}$ from its corresponding output node feature $v_i^n$ (i.e., decoding an appropriate facial reaction from the predicted distribution), which is formulated as:
\begin{equation}
\label{eq:inverse}
\begin{aligned}
     & v^{n-1}_i = \text{INorm}(x_{k}) ~~~~~~  \textbf{subject to} \\
     & x_{k}  = v^{n}_i - \varphi(x_{k-1}) ~~  \text{and} ~~  x_{k} = x_{k-1}
\end{aligned}
\end{equation}
where INorm denotes the inverse normalisation; the function $\varphi$ is defined in Eq. \ref{eq:phi}; \textbf{$x_{k}$ is computed iteratively until it is converged to $x_{k} = x_{k-1}$, and consequently $v^{n-1}_i = x_k$ (please refer to Algorithm \ref{algorithm: 2} for details of this iterative process)}. Here, the $x_{0}$ can be set as a non-zero random value. To achieve the aforementioned reversibility (i.e., converged at the $x_{k} = x_{k-1}$), the function $\varphi$ needs to be a contraction mapping \cite{lipattn}. Consequently, we further define the function $\varphi$ as:
\begin{equation}
\label{eq:final updating}
    \varphi(v_i^{n-1}) = \frac{g(\text{Sig}(v_i^{n-1}))}{1 + 2\|\ \*{W}_q^{n} {\*{W}_m^{n}}^{\top}\|_2},
\end{equation}
where the function $g$ is defined as the combination of the Edge Update (Eq.~\ref{eq: edge updating}) and Message Passing functions (Eq.\ref{eq: message passing}) involved in the forward propagation, and $\text{Sig}$ denotes the Sigmoid activation function. Particularly, the use of the Sigmoid function and the term $1 + 2\|\ \*{W}_q^{n} {\*{W}_m^{n}}^{\top}\|_2$ ensures the function $\varphi$ to be a contraction mapping function (i.e., the function is Lipschitz continuous and its Lipschitz constant less than $1$ \cite{lipattn, fixedpoint}). Consequently, there exists a fixed point $x_{k}$ satisfying the equation $x=v_i^n - \varphi(x)$ (i.e., achieving Eq. \ref{eq:inverse}). The detailed proof and derivation for defining Eq. \eqref{eq:final updating} for the reversibility of the proposed REGNN are provided in the supplementary material.

\section{Experiments}

\noindent This section first provides the details of experimental settings in Sec. \ref{subsec:experimental setting}. Then, Sec. \ref{subsec:comparison to baselines} comprehensively compares our approach with previous facial reaction generation solutions. Finally, we conduct a series of ablation studies and parameter sensitivity analysis to investigate the contributions of different modules (Sec. \ref{subsec: ablation}), as well as the robustness (Sec. \ref{subsec: sensativity}) of the proposed approach.

\begin{figure*}[tp]
	\begin{center}
		\includegraphics[width=1\textwidth]{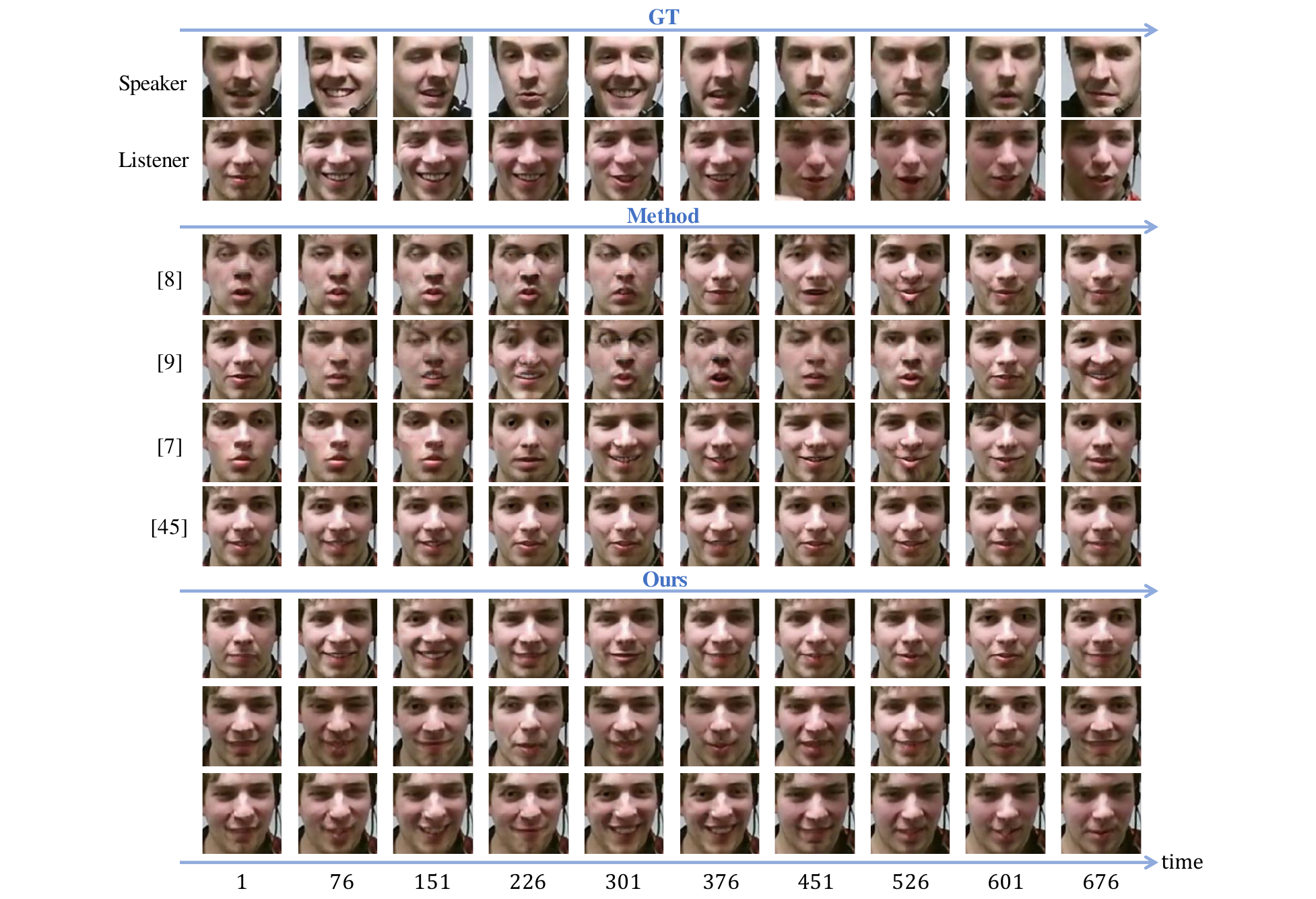}
	\end{center}
	\caption{Visualisation of the facial reactions generated from different approaches, where early approaches \cite{CGAN-BMVC,CGAN-CVPR,unet} generated some very low-quality facial images, while the predictions of a recent approach \cite{ng2022learning} is quite different from the ground-truth (i.e., low appropriateness and synchrony). Our approach generated multiple diverse but appropriate, realistic, and synchronized facial reactions from the input speaker behaviour.}
	\label{fig:visual}
\end{figure*}

\subsection{Experimental settings}
\label{subsec:experimental setting}

\subsubsection{Dataset} 

\noindent This paper evaluates the proposed approach based on video conference clip pairs recorded under various dyadic interaction settings, which are provided by two publicly available datasets: NoXI \cite{cafaro2017noxi} and RECOLA \cite{ringeval2013introducing}. As there are only three valid video pairs in RECOLA dataset, experiments can not be individually conducted on this dataset. Thus, we follow a public facial reaction challenge \footnote{https://sites.google.com/cam.ac.uk/react2023/home} to combine and split these two datasets, resulting in 2962 pairs of audio-visual speaker-listener dyadic interaction clips. This includes 1594 pairs of training clips, 562 pairs of validation clips, and 806 pairs of test clips, where each clip is 30 seconds long. All appropriateness labels (i.e., the appropriate real facial reactions for each speaker behaviour) used in our experiments are provided by the challenge, which are obtained based on the automatic and objective labelling strategy proposed by \cite{song2023multiple}. Note that the employed dataset is different from the challenge, as the UDIVA dataset \cite{palmero2021context} is not included in our experiments because it is recorded under in-person dyadic interactions (with a lateral camera view which captures both participants), where not only the profile of participants' faces are frequently recorded but also the conversational partners' faces are sometimes incorrectly recorded.



\subsubsection{Implementation details}

\noindent In our experiments, the input of the \textbf{Perceptual Processor} includes the full facial image sequence cropped (i.e., OpenFace 2.0 \cite{baltrusaitis2018openface} is employed) from each 30s speaker behaviour video, and 128 dimensional log-mel spectrogram features extracted from the corresponding 30s speaker behaviour audio. At the training stage, we employ the Swin-Transformer (i.e., the tiny version) \cite{liu2021swinv2} pre-trained on FER2013 \cite{song2022gratis}, and pre-trained VGGish model provided by \cite{hershey2017cnn} as the initial facial and audio feature extraction encoder (i.e., $\textbf{Enc}_\textbf{F}$ and $\textbf{Enc}_\textbf{A}$), while the REGNN used in our experiments consists of six layers. Then, the Adam optimizer \cite{kingma2014adam} was employed to train the entire framework in an end-to-end manner using an initial learning rate of $10^{-4}$ and weight decay of $5 \times 10^{-4}$. The maximum number of epochs was set to $100$, with learning rate decay ($0.1$) performed at the $20^{th}$ and $50^{th}$ epochs. We empirically set the $\sigma = 0.6$ for all GMM in the GMGD, based on which the REGNN reversely samples and decodes facial reactions at the inference stage. The model proposed in \cite{ganimation} is finally employed to generate facial images from all predicted AUs, where we follow the same train strategy to re-train it based on 15 automatically detected AUs and corresponding facial images contained in our training set.

\subsubsection{Evaluation metrics} 

\noindent We employed the four sets of metrics defined in \cite{song2023multiple} to evaluate four perspectives of the generated facial reactions, in terms of: (i) \textbf{Appropriateness}: the distance and correlation between generated facial reactions and their most similar appropriate facial reaction using \textbf{FRDist} and \textbf{FRCorr}. In addition, we also report the Pearson Correlation Coefficient (PCC) between predictions and their most similar appropriate real facial reaction; (ii) \textbf{Diversity}: the variation among 1) frames in each generated facial reaction (\textbf{FRVar}), 2) multiple facial reactions generated from the same speaker behaviour (\textbf{FRDiv}), and 3) facial reactions generated for different speaker behaviours (\textbf{FRDvs}); (iii) \textbf{Realism}: employing Fréchet Inception Distance (FID) used in \cite{song2023multiple} (\textbf{FRRea}); and (iv) \textbf{Synchrony} is measured by Time Lagged Cross Correlation (TLCC) between the speaker facial behaviour and the corresponding generated facial reaction.

\subsection{Comparison to related works}
\label{subsec:comparison to baselines}

\begin{table*}
    \normalsize
    \begin{center}
    \caption{Comparison between the proposed approach and several reproduced existing related works.}
    \label{tab:baselines}
    \begin{tabular}{lrrrrrrrrrr}
    \toprule
    Methods & FRDist $\downarrow$ & FRCorr $\uparrow$ & PCC $\uparrow$ & FRRea $\downarrow$ & FRVar $\uparrow$ & FRDvs $\uparrow$ & FRDiv $\uparrow$ & Synchrony $\downarrow$ \\
    \midrule
    GT & 0.00 & 0.873 & 0.989 & - & 0.072 & 0.248 & 0.000 & 47.69\\
    \midrule
    B\_Random & 237.21 & 0.005 & 0.007 & - & 0.083 & 0.167 & 0.167 & 44.19 \\
    B\_Mime & 92.94 & 0.038 & 0.058 & - & 0.072 & 0.248 & 0.000 & 38.54 \\
    B\_MeanSeq & 97.12 & 0.001 & 0.001 & - & 0.000 & 0.000 & 0.000 & 45.27 \\ 
    B\_MeanFr & 97.85 & 0.000 & 0.000 & - & 0.000 & 0.000 & 0.000 & 49.00 \\ \hline
    Huang et al. (S) \cite{CGAN-BMVC} & 14.91 & 0.023 & 0.045 & 42.44 & \textbf{0.092} & \textbf{0.186} & \textbf{0.181} & 42.83\\
    Huang et al. (C) \cite{CGAN-CVPR} & 11.99 & 0.035 & 0.058 & 32.53 & 0.054 & 0.114 & 0.102 & 43.83\\
    UNet \cite{unet}   & 8.31  & 0.067 & 0.141 & 23.81 & 0.015 & 0.049 & 0.000 & 40.19\\
    Ng et al. \cite{ng2022learning} & 10.36 & 0.082 & 0.146 & 27.33 & 0.075 & 0.064 & 0.061 & 39.22 \\ \hline
    \rowcolor{Gray}
    Ours    & \textbf{7.62}  & \textbf{0.106} & \textbf{0.153} & \textbf{21.58}  & 0.077 & 0.121 & 0.048 & \textbf{38.76}\\
    \bottomrule
    \end{tabular}
    \end{center}
\end{table*}

\begin{figure*}[tp]
	\begin{center}
		\includegraphics[width=0.96\hsize]{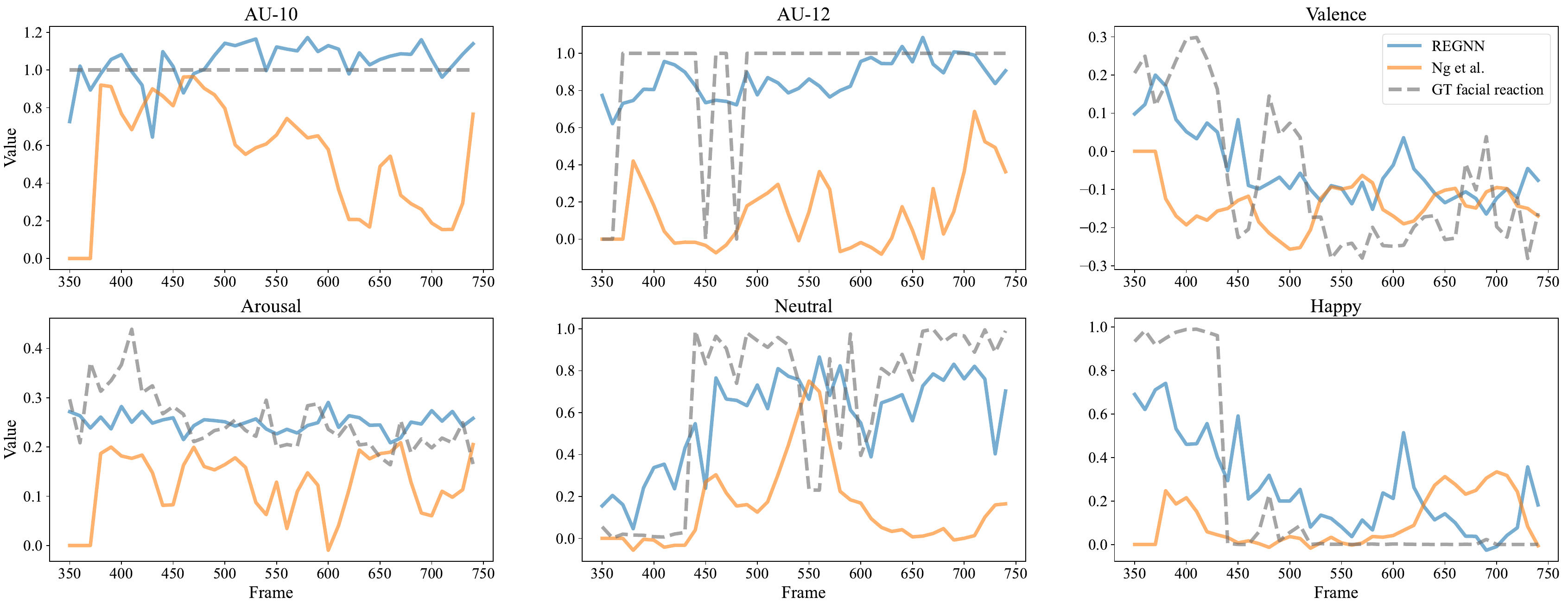}
	\end{center}
	\caption{Visualisation of the facial attribute predictions. Although our approach is trained to generate appropriate facial reactions, the facial attributes predicted by our approach are still highly correlated with the GT real facial reaction.}
	\label{fig:timeseries}
\end{figure*}

\noindent As this paper presents the first approach aiming to generate multiple appropriate facial reactions, we compare it with different reproduced baselines \cite{CGAN-BMVC,CGAN-CVPR,unet,ng2022learning} that have been previously used for generating facial reactions in Table \ref{tab:baselines} (the details of these reproduced approaches are provided in the supplementary material). It can be observed that our approach outperforms all competitors in generating appropriate, realistic, and synchronized facial reactions, as indicated by the lowest FRDist/FRCorr/PCC distances (appropriateness), FID (Realism), and TLCC (Synchrony) values between the facial reactions generated by our approach and their most similar appropriate real facial reactions. Specifically, our approach achieved $4.37$ and $2.74$ absolute improvements in FRDist, $0.071$ and $0.024$ absolute improvements in FRCorr, as well as $0.095$ and $0.007$ absolute improvements in PCC over the condition GAN-based approach \cite{CGAN-CVPR} and the recently proposed VQ-VAE based approach \cite{ng2022learning}, respectively. Furthermore, our approach generates diverse facial reactions in response to different speaker behaviours, as well as decent diversity among frames of each generated facial reaction. As visualised in Fig. \ref{fig:visual}, the three facial reaction sequences generated by our approach in response to the same example speaker behaviour are diverse, where all of them show positive emotions but displayed by different facial behaviours (e.g., $76_\text{th}$ and $301_\text{th}$ frames). In addition, Fig. \ref{fig:timeseries} specifically compare the facial reaction attributes predicted by our approach with these predicted by the VQ-VAE approach proposed in \cite{ng2022learning}. It is clear that the facial reaction attributes generated by our approach are more correlated with the corresponding ground-truth real facial reactions expressed by human listeners.

\begin{figure}[tp]
	\begin{center}
		\includegraphics[width=1\hsize]{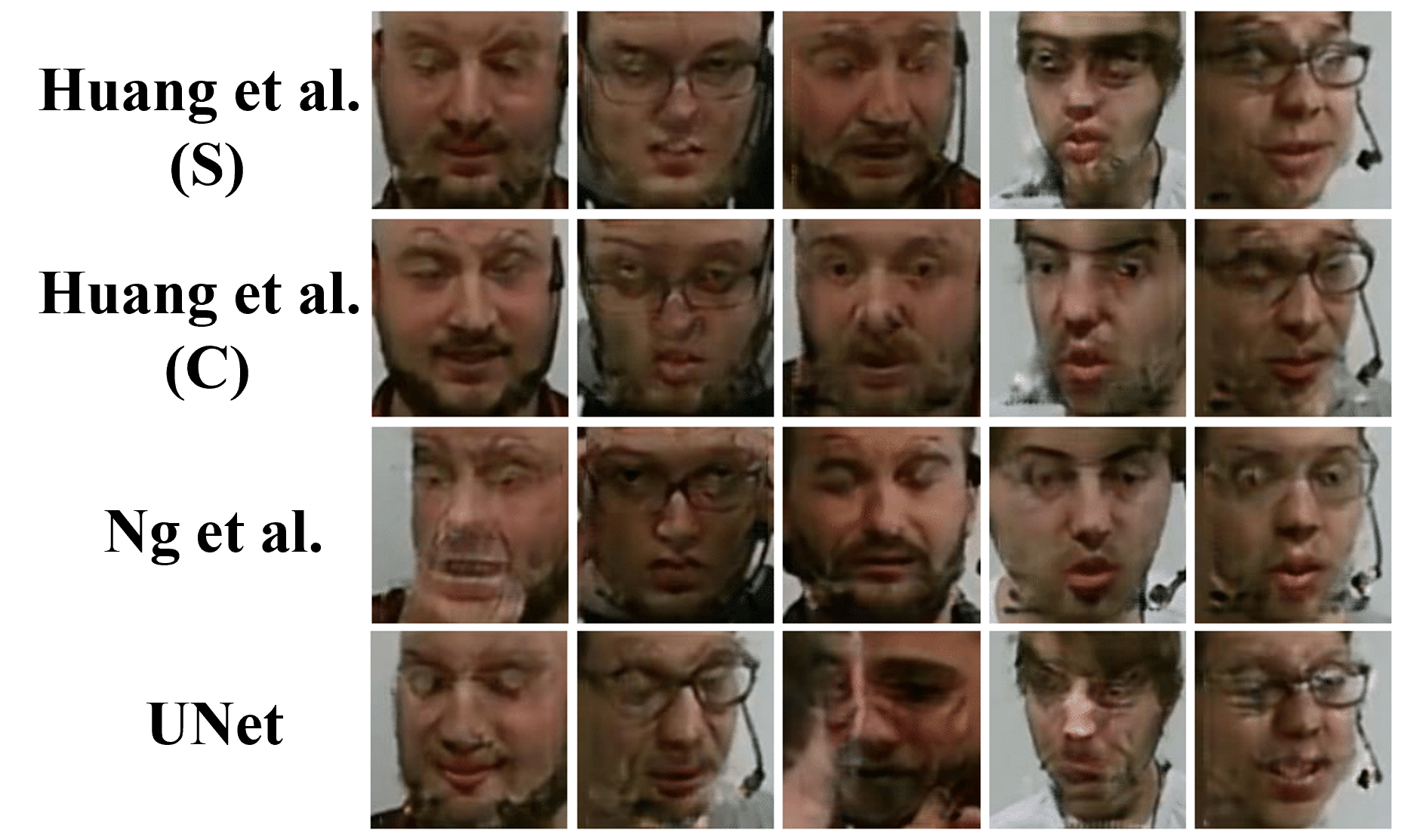}
	\end{center}
	\caption{Examples of the low-quality/abnormal facial reaction images generated from the facial reaction attributes predicted by competitors.}
	\label{fig:visual2}
\end{figure}

\begin{figure*}[tp]
	\begin{center}
		\includegraphics[width=0.916\hsize]{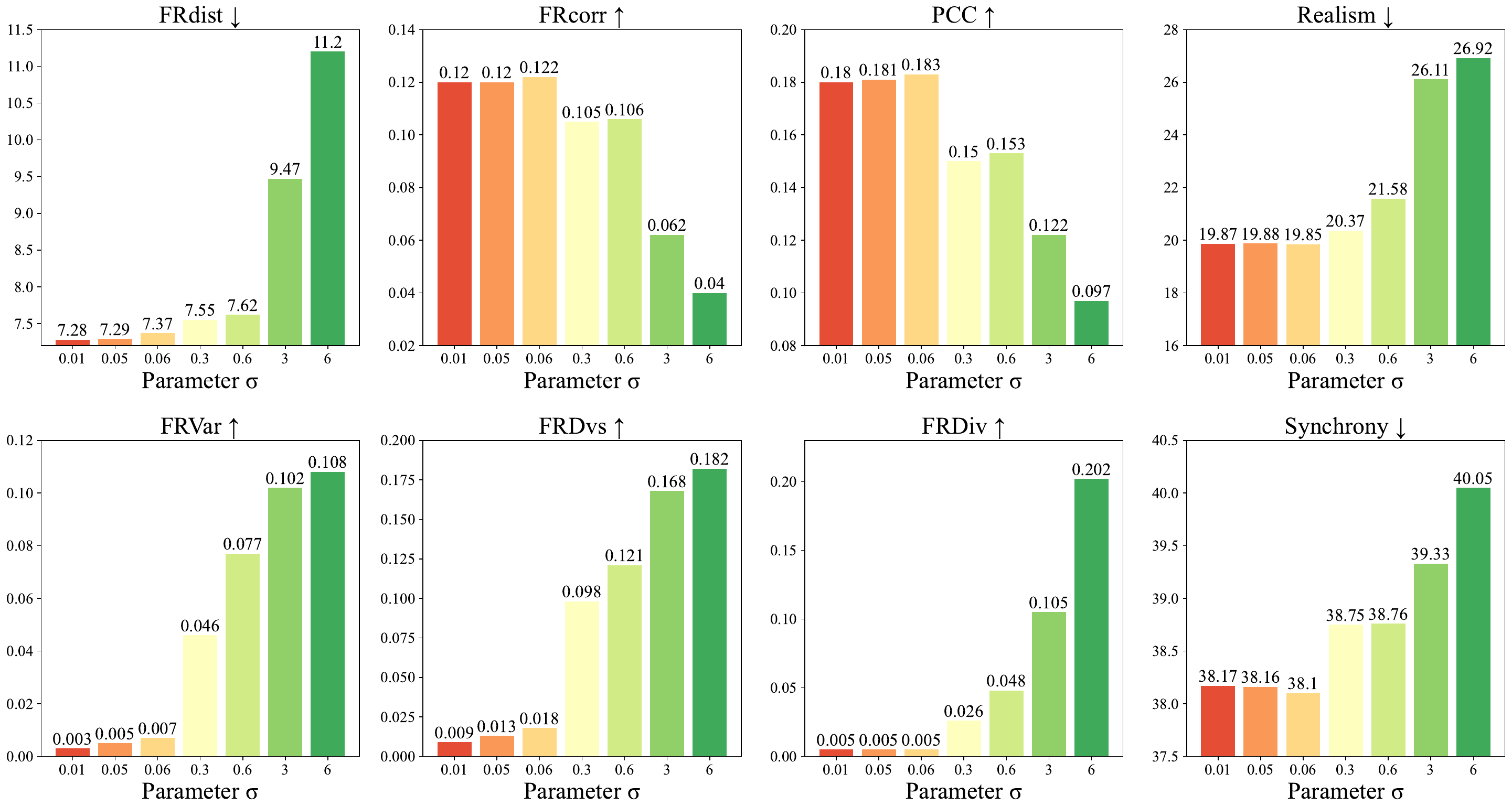}
	\end{center}
	\caption{Impacts of different $\sigma$ settings on the facial reaction generation performances.}
	\label{fig:param}
\end{figure*}

These results discussed above demonstrate the effectiveness of our approach in generating multiple different but appropriate, realistic, and synchronized facial reactions. It should be noted that while our approach did not generate facial reactions with as much diversity as C-GAN based \cite{CGAN-BMVC,CGAN-CVPR} and VQ-VAE \cite{ng2022learning} approaches, their high diversity results are partially associated to the generation of abnormal facial reactions (i.e., facial behaviours that are not appropriate or cannot be properly expressed by humans (illustrated in Fig.\ref{fig:visual2})), which is reflected by their much worse appropriateness, realism, and synchrony performances.

\begin{figure}[tp]
	\begin{center}
		\includegraphics[width=0.96\hsize]{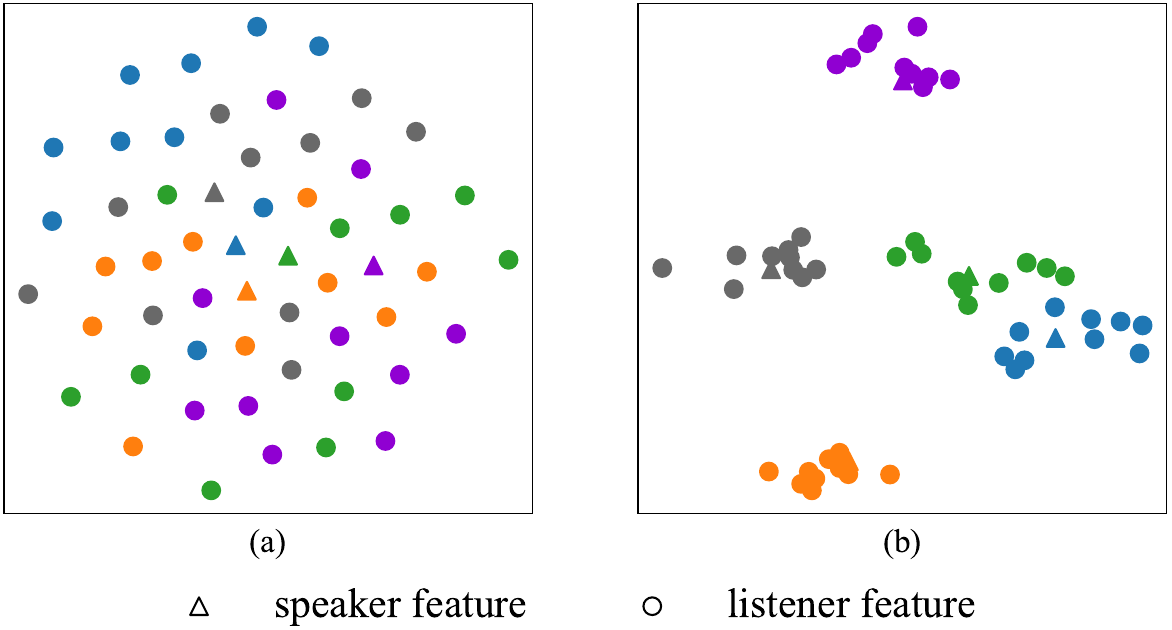}
	\end{center}
	\caption{Visualisation of the learned distributions. It is clear that the distribution learned by the proposed REGNN (depicted in (b)) are more discriminative than i-ResNet (depicted in (a)).}
	\label{fig:visual_dis}
\end{figure}

\subsection{Ablation studies}
\label{subsec: ablation}

\noindent In this section, we first conduct a series of ablation studies to evaluate the effectiveness/importance of (i) each modality of the speaker behaviour; (ii) the proposed appropriate facial reaction distribution learning (AFRDL) strategy; (iii) the proposed reversible graph model (REGNN); and (iv) the multi-dimensional edge feature learning (MEFL) module.

\textbf{Contributions of different modalities.} The experimental results reported in Table \ref{tab:ablation} reveal that both audio and facial modalities of the speaker behaviour offer valuable cues for generating appropriate facial reactions, as the facial reactions individually generated by each modality show a positive correlation with the corresponding real appropriate facial reactions. Particularly, speakers' facial behaviours provided greater contributions to the generated facial reactions, with clearly better performances in terms of appropriateness, diversity, and realism. Since the fusion of speaker audio and facial behaviours results in the best performance, it suggests that audio and visual cues from the speaker behaviour are complementary and relevant for the generation of appropriate facial reactions. Importantly, the proposed approach outperforms several existing methods \cite{CGAN-BMVC,CGAN-CVPR,ng2022learning} even when either audio or visual modality of the speaker behaviour alone is utilized, which further validated the effectiveness of the proposed approach.

\begin{table*}
    \begin{center}
    \caption{Results achieved for four ablation studies.}
    \label{tab:ablation}
    \begin{tabular}{llrrrrrrrrrr}
    \toprule
    & Methods & FRDist $\downarrow$ & FRCorr $\uparrow$ & PCC $\uparrow$ & FRRea $\downarrow$ & FRVar $\uparrow$ & FRDvs $\uparrow$ & FRDiv $\uparrow$ & Synchrony $\downarrow$ \\
    \midrule
    \multirow{3}{*}{Modalities} & Video & 8.33 & 0.101 & 0.129 & 24.97 & 0.013 & 0.026 & 0.052 & 38.62\\
    & Audio  & 8.81 & 0.031 & 0.062 & 30.54 & 0.023 & 0.047 & \textbf{0.081} & 40.33\\
    & \cellcolor{Gray}Audio+Video   & \cellcolor{Gray} \textbf{7.62}  
    & \cellcolor{Gray} \textbf{0.106} & \cellcolor{Gray} \textbf{0.153} 
    & \cellcolor{Gray} \textbf{21.58}  & \cellcolor{Gray} \textbf{0.077} 
    & \cellcolor{Gray} \textbf{0.121} & \cellcolor{Gray} 0.048 
    & \cellcolor{Gray} \textbf{38.76}\\
    \midrule
    \multirow{2}{*}{AFRDL} & Without Graph  & 11.6 & 0.048 & 0.070 & 29.71 & 0.075 & \textbf{0.128} & \textbf{0.076} & 42.43\\
    & \cellcolor{Gray}With AFRDL & \cellcolor{Gray} \textbf{7.62}  
    & \cellcolor{Gray} \textbf{0.106} & \cellcolor{Gray} \textbf{0.153} 
    & \cellcolor{Gray} \textbf{21.58} & \cellcolor{Gray} \textbf{0.077} 
    & \cellcolor{Gray} 0.121 & \cellcolor{Gray} 0.048 
    & \cellcolor{Gray} 38.76\\
    \midrule
    \multirow{2}{*}{Motor processor} & i-ResNet \cite{iresnet} & 15.52 & 0.051 & 0.077 & 35.20  & \textbf{0.090} & \textbf{0.193} & \textbf{0.169} & 42.12\\
    & \cellcolor{Gray}REGNN & \cellcolor{Gray} \textbf{7.62}  
    & \cellcolor{Gray} \textbf{0.106} & \cellcolor{Gray} \textbf{0.153} 
    & \cellcolor{Gray} \textbf{21.58} & \cellcolor{Gray} 0.077 
    & \cellcolor{Gray} 0.121 & \cellcolor{Gray} 0.048 
    & \cellcolor{Gray} \textbf{38.76}\\
    \midrule
    \multirow{2}{*}{Edge feature} & Without MEFL  & 8.97 & 0.045 & 0.096 & \textbf{20.33} & 0.031 & 0.110 & \textbf{0.062} & 38.91 \\
    & \cellcolor{Gray}With MEFL & \cellcolor{Gray}\textbf{7.62}  
    & \cellcolor{Gray}\textbf{0.106} & \cellcolor{Gray}\textbf{0.153} 
    & \cellcolor{Gray}21.58 & \cellcolor{Gray}\textbf{0.077} 
    & \cellcolor{Gray}\textbf{0.121} & \cellcolor{Gray}0.048 
    & \cellcolor{Gray}\textbf{38.76}\\
    \bottomrule
    \end{tabular}
    \end{center}
\end{table*}

\textbf{Appropriate facial reaction distribution learning (AFRDL) strategy.} Table \ref{tab:ablation} reports a comparative analysis for the effectiveness of the proposed AFRDL strategy, where we developed a variant of our framework with the same architecture but different training strategy. The variant was trained using MSE loss, which minimizes the difference between the generated facial reaction and the GT real facial reaction (the specific real facial reaction expressed by the listener in response to the input speaker behaviour). The achieved results indicate that the proposed AFRDL strategy is crucial for giving model the ability to generate high-quality facial reactions, as the variant of the same architecture but different training strategy achieved much worse performance in terms of appropriateness, realism, and synchrony. Moreover, this also suggests that the necessarily of the proposed reversible GNN network which is the key for achieving the AFRDL strategy.

\textbf{REGNN vs. Reversible CNN.} Consequently, we also compare performances achieved our reversible GNN (REGNN) with a widely-used reversible CNN (i-ResNet \cite{iresnet}) in Table \ref{tab:ablation} and Fig. \ref{fig:visual_dis}, to show the superiority of the proposed REGNN. Except the network employed for the motor processor, the rest of the framework and training strategy are kept the same for both experiments. The achieved results demonstrate that the REGNN-based systems (i.e., both single-value edge graph-based system and multi-dimensional edge graph-based system) outperformed the i-ResNet-based system with substantial improvements across all metrics of appropriateness, realism, and synchrony.  As discussed in Sec. \ref{subsec:comparison to baselines}, the i-ResNet-based system sometimes generates abnormal facial reactions, which may lead to its better diversity performance (illustrated in Fig. \ref{fig:visual2}). In other words, the proposed REGNN allows the framework to generate more appropriate, realistic and synchronised facial reactions over the Reversible CNN. We hypothesis that this is because the REGNN can explicitly represents the task-specific relationship between each pair of facial attributes in the form of multi-dimensional edge features (i.e., this can not be achieved by Reversible CNN), which could be crucial for predicting facial reactions.

\textbf{The MEFL module.} Finally, it can be seen that the multi-dimensional edge features generated by the proposed MEFL module also enhanced the quality of the generated facial reactions. As shown in Table \ref{tab:ablation}, the additional usage of the MEFL module provides large improvements in terms of all appropriateness metrics (i.e., $15\%, 135\%$, and $59\%$ relative improvements in FRDist, FRCorr, and PCC, respectively) as well as two diversity metrics (i.e., FRVar and FRDvs), showing that the task-specific relationships between facial attributes could be complex, and thus were better modeled by multi-dimensional edge features rather than single-value edge features. In contrast, the multi-dimensional edge features learned by our MEFL module only have small impacts on the generated facial reactions' realism and synchrony performances.

\subsection{Parameter sensitivity analysis}
\label{subsec: sensativity}

\noindent We provide the sensitivity analysis for three main parameters: (i) the $\sigma$ used for defining the proposed Gaussian Mixture Graph Distribution (GMGD) $Z_L(B_S^{t_1,t_2})$; (ii) the dimension $D$ of multi-dimensional edge features generated from the MEFL module; and (iii) the number of employed REGNN layers in the Motor Processor.

\textbf{Sensitivity of the parameter $\sigma$:} Fig. \ref{fig:param} evaluates the sensitivity of the $\sigma$ used for defining the Gaussian Mixture Graph Distribution (GMGD) $Z_L(B_S^{t_1,t_2})$. It can be observed that there is a clear trade-off between appropriateness performances and diversity performances, i.e., with the increasing of the $\sigma$, the appropriateness performances are degraded while the diversity performances are increased. However, we found that when $\sigma < 0.1$, the appropriateness, realism and synchrony performances are relatively stable and promising (i.e., they are roughly convergent at $\sigma \approx 0.6$.).

\begin{figure*}[tp]
	\begin{center}
		\includegraphics[width=0.916\hsize]{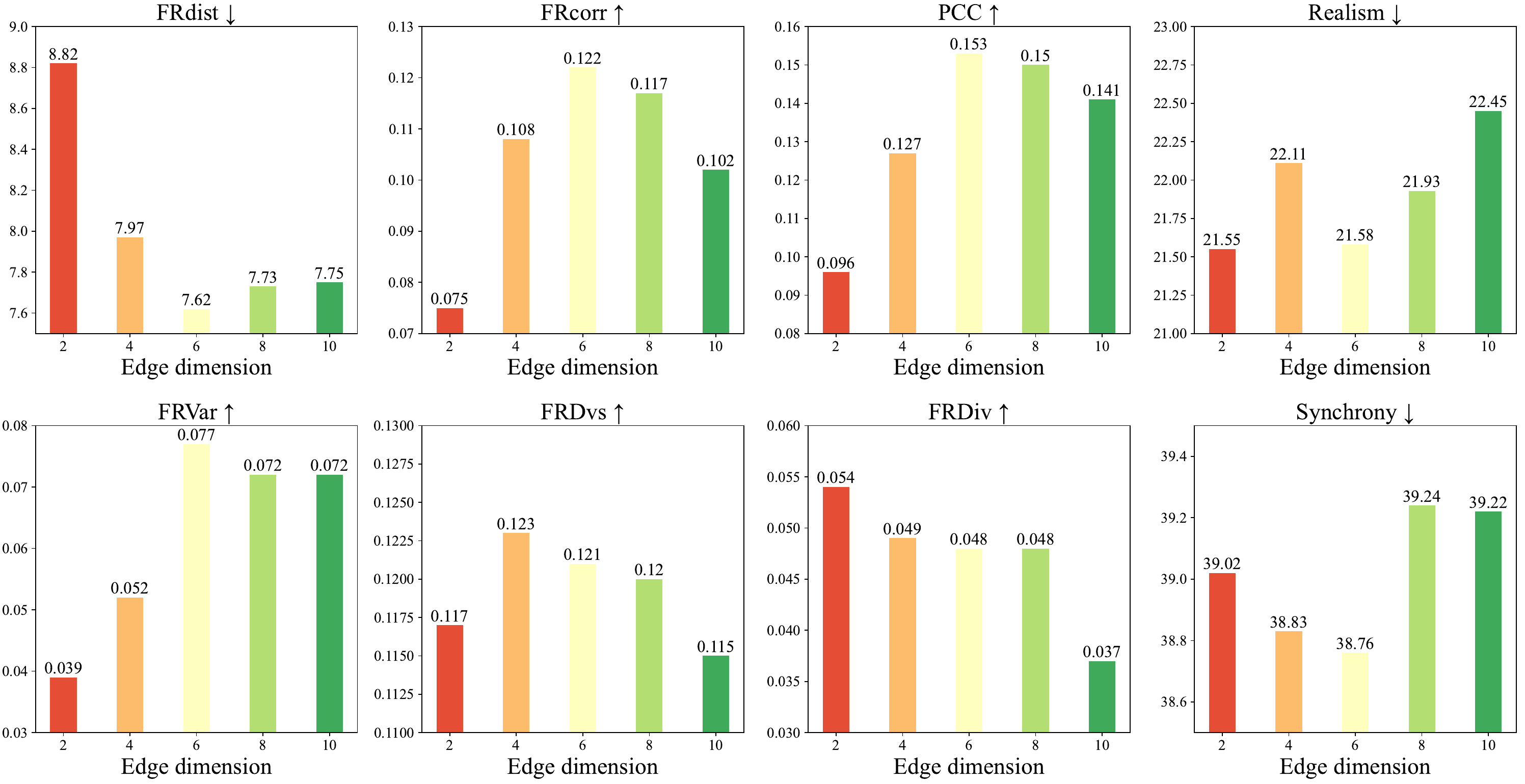}
	\end{center}
	\caption{Impacts of edge dimension settings on the facial reaction generation performances.}
	\label{fig:edge-dim}
\end{figure*}

\textbf{Edge feature dimension:} Fig. \ref{fig:edge-dim} also evaluates the impact of the dimension for the generated multi-dimensional edge features. While the performances are fluctuated with the change of the edge dimension, the performances of some metrics are relatively stable (i.e., the variations are relatively small for FRDvs, Realism and Synchrony metrics). It also can be observed that most metrics have the best performances when edge dimension is set to $D = 6$.

\textbf{The number of employed REGNN layers:} Finally, we found that the performances of the REGNN is relatively robust to the number of employed layers. Specifically, when the number of layers ranges from 4 to 10, the CCC and PCC results only changed less than $0.02$, while FRDvs and FRDiv only changed less than $0.01$. In addition, the relatively changes in Realism and Synchrony metrics are even less than $4.1\%$ and $1.2\%$, respectively. Importantly, we found when the number of REGNN layers is 6, the proposed approach achieved the best performances in five out of eight metrics, while produced the second best results on FRDvs and Realism metrics.

\begin{figure*}[tp]
	\begin{center}
		\includegraphics[width=0.916\hsize]{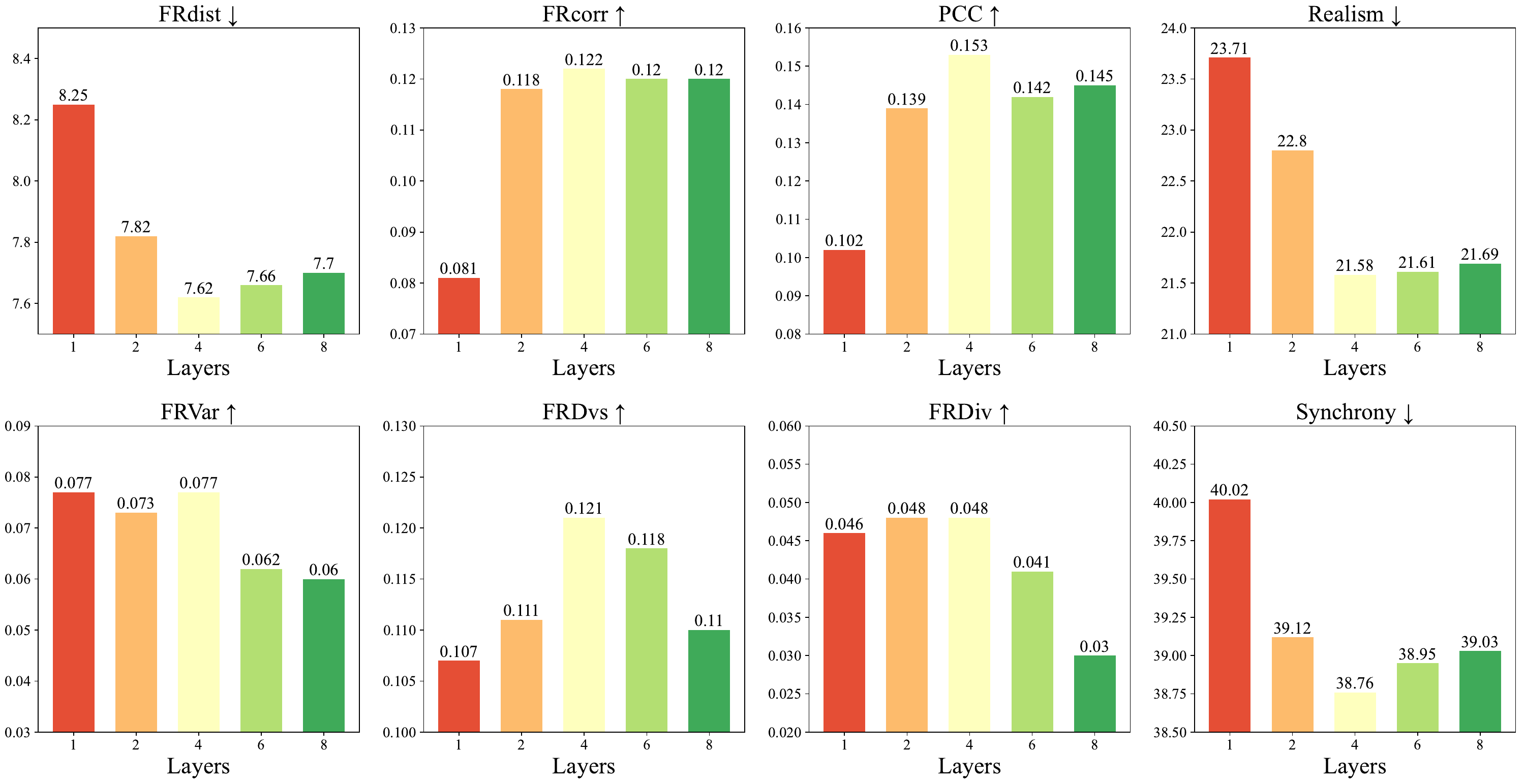}
	\end{center}
	\caption{Impacts of the number of REGNN layers on the facial reaction generation performances.}
	\label{fig:layers}
\end{figure*}

\section{Conclusion}

\noindent In this paper, we propose the first automatic multiple appropriate human facial reaction generation deep learning framework, which opens up a new avenue of research for predicting multiple appropriate human facial reactions in response to each speaker behaviour. Importantly, we propose the first solution that reformulates the ``one-to-many mapping'' problem occurring in training FRG models as a ‘one-to-one mapping’ problem (one speaker behaviour corresponding to one distribution representing multiple appropriate facial reactions), where a novel reversible GNN (REGNN) and a novel multiple appropriate facial reaction distribution learning (AFRDL) strategy are proposed.

As the first specifically designed multiple appropriate FRG model, we compared our approach with a set of reproduced baselines. The experimental results show that: (i) our approach can learn useful cues from speaker behaviours for predicting semantic meaningful human-style facial reactions, as it achieved large performance gains compared to four basic baselines; (ii) our approach can generate multiple diverse but appropriate, realistic, and synchronized facial reactions in response to each speaker behaviour, and achieve greater performance in appropriateness, realism, and synchrony metrics as compared to all the reproduced existing FRG approaches; (iii) the proposed REGNN-based facial reaction distribution learning contributes substantially to the promising appropriateness, realism, and synchrony performances achieved by our approach, where the number of REGNN layers; (iv) both audio and facial speaker behaviours provide relevant and complementary information; (v) the proposed REGNN is crucial for the success of the AFRDL strategy; (vi) the MEFL module is crucial for generating appropriate facial reactions, as multi-dimensional edge features generated by it can comprehensively model task-specific relationships among facial attributes.

\textbf{Limitations and future work:} As the first automatic multiple appropriate FRG model, this paper only predicted facial reactions based on non-verbal behaviours expressed by speakers while ignoring important verbal textual cues. Another limitation is that sometimes the appropriate facial reaction distribution for different speaker behaviours are similar, which may negatively affect the model's training process. Both limitations may lead to the limited performances achieved by the proposed approach. Finally, due to the limited resources, we are not able to reproduce all generative deep learning approaches (e.g., different GANs and diffusion models) for the MAFRG task but only reproduced several approaches that have been already proposed for facial reaction generation. As a result, our future work will focus on (i) developing more advanced MAFRG-specific generative models; (ii) considering both verbal and non-verbal behaviours of speakers; and (iii) investigating better ways to represent appropriate facial reaction distributions.


\ifCLASSOPTIONcaptionsoff
  \newpage
\fi

{\small
	\bibliographystyle{IEEEtran}
	\bibliography{egbib}
}

\begin{IEEEbiography}
[{\includegraphics[width=1in,height=1.25in,clip,keepaspectratio]{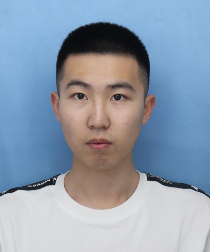}}]{Tong Xu} is a researcher at JD Inc, and this work was done during his internship at the University of Leicester, UK. His main research interests are multimodal understanding and 3D vision.
\end{IEEEbiography}

\begin{IEEEbiography}[{\includegraphics[width=1in,height=1.25in,clip,keepaspectratio]{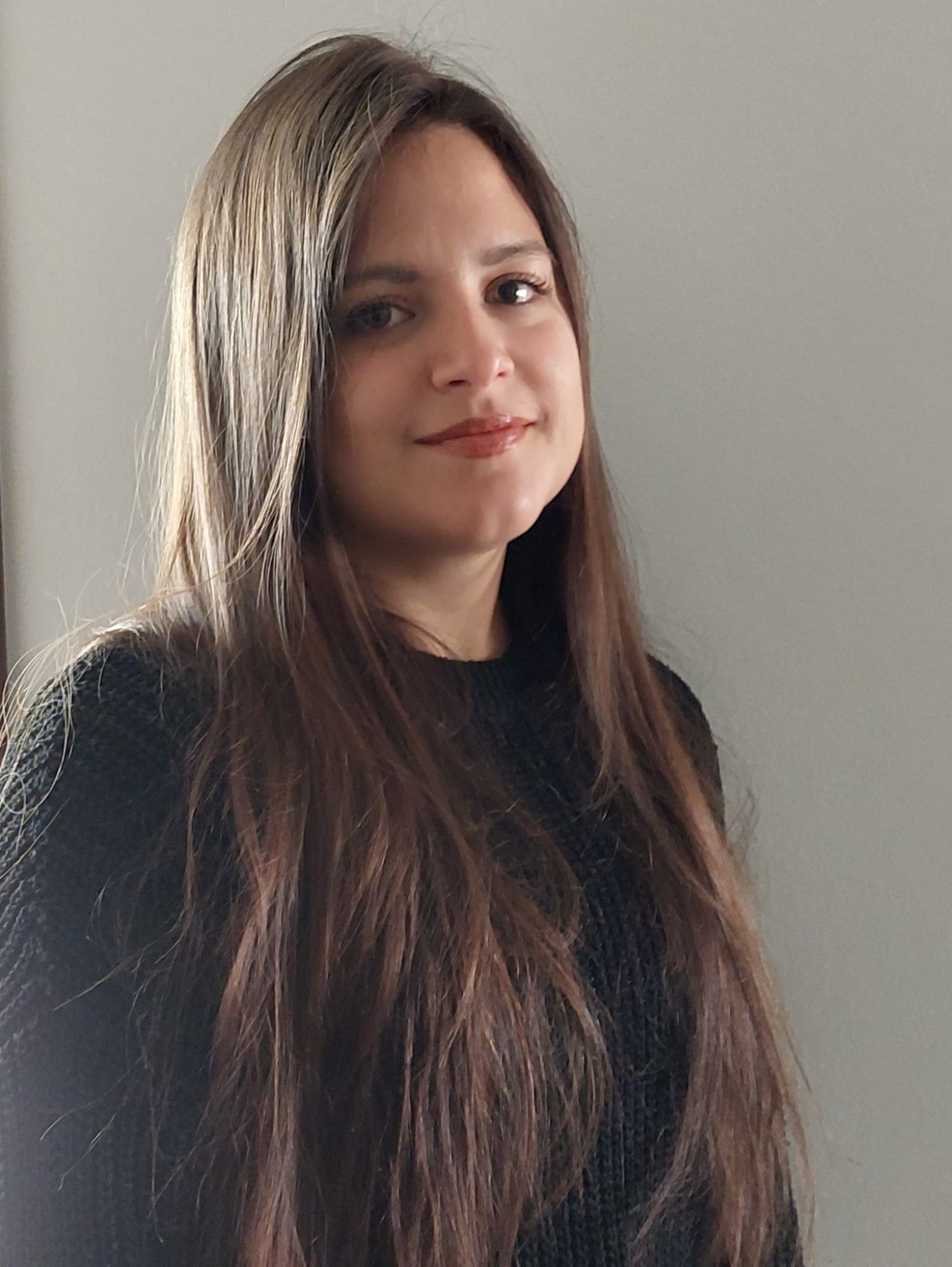}}]{Micol Spitale} is currently a PostDoctoral Researcher at the Affective Intelligence \& Robotics Laboratory (AFAR Lab), Department of Computer Science \& Technology, of the University of Cambridge, UK under the supervision of Prof. Hatice Gunes. Her research activities are grounded in the Social Robotics area. She has a strong background in affective computing, child-robot interaction, and machine learning applications to human behavioural analysis. Her current research focuses on developing socio-emotionally adaptive robots that can foster wellbeing through coaching and psychologically proven interventions. She has been awarded “cum laude” a Ph.D. in Information Technology, Computer Science and Engineering Area at the Politecnico di Milano, co-funded by IBM Italy and EIT Digital, in October 2021.
\end{IEEEbiography}

\begin{IEEEbiography}[{\includegraphics[width=1in,height=1.25in,clip,keepaspectratio]{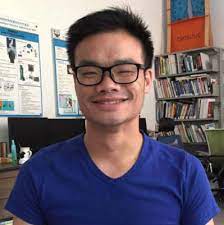}}]{Hao Tang} received the master’s degree from the School of Electronics and Computer Engineering, Peking University, China, and the PhD degree from the Multimedia and Human Understanding Group, University of Trento, Italy. He is currently a postdoctoral with Computer Vision Lab, ETH Zurich, Switzerland. He was a visiting scholar with the Department of Engineering Science, University of Oxford. His research interests include deep learning, machine learning, and their applications to computer vision.
\end{IEEEbiography}

\begin{IEEEbiography}[{\includegraphics[width=1in,height=1.25in,clip,keepaspectratio]{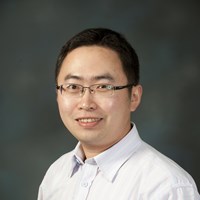}}]{Lu Liu} is a Professor and Head of the School of Computing and Mathematical Sciences at the University of Leicester, UK. Prof. Liu received his Ph.D. degree from the University of Surrey and M.Sc. degree from Brunel University. Prof. Liu’s research interests are in the areas of data analytics, sustainable computing, service computing, artificial intelligence and the Internet of Things. He is a Fellow of British Computer Society (BCS).
\end{IEEEbiography}

\begin{IEEEbiography}[{\includegraphics[width=1in,height=1.25in,clip,keepaspectratio]{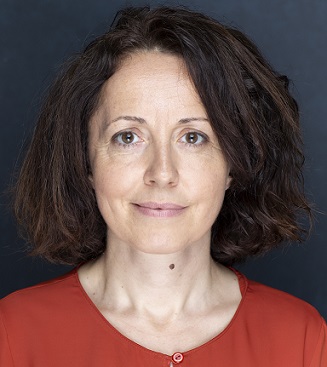}}]{Hatice Gunes} is a Professor with the Department of Computer Science and Technology, University of Cambridge, U.K., leading the Affective Intelligence and Robotics Lab. Her expertise is in the areas of affective computing and social signal processing cross-fertilizing research in human behaviour understanding, computer vision, machine learning, and human–robot interaction. She has published over 150 papers in the above areas, and her research highlights  include  RSJ/KROS  Distinguished  Interdisciplinary Research Award Finalist at IEEE RO-MAN’21, Distinguished PC  Award  at  IJCAI’21,  Best  Paper  Award  Finalist  at  IEEE RO-MAN’20, Finalist for the 2018 Frontiers Spotlight Award, Outstanding  Paper  Award  at  IEEE  FG’11,  and  Best  Demo Award at IEEE ACII’09. Prof Gunes is a former President of the Association for the Advancement of Affective Computing (AAAC). Her research has been supported by various competitive grants, with funding from the Engineering and Physical Sciences Research Council, UK (EPSRC), Innovate UK, British Council, Alan Turing Institute and EU Horizon 2020. She is a Fellow of the EPSRC, a Staff Fellow of Trinity Hall Cambridge, and was a Faculty Fellow of the Alan Turing Institute.
\end{IEEEbiography}

\begin{IEEEbiography}
[{\includegraphics[width=1in,height=1.25in,clip,keepaspectratio]{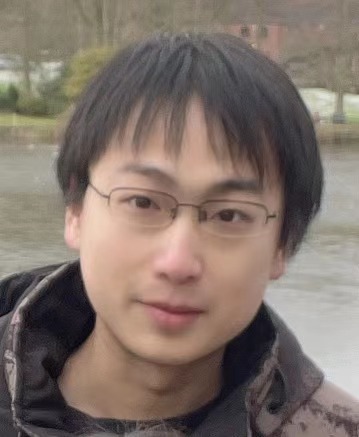}}]{Siyang Song} is a Lecturer (Assistant Professor) at the University of Leicester, UK. He is also an affiliated/visiting researcher at the Department of Computer Science and Technology, UK, University of Cambridge. He received his PhD in the Computer Vision Lab and Horizon Center for Doctoral Training (CDT) at the University of Nottingham, UK. His research interests include automatic facial reaction generation as well as emotion, personality and depression analysis by developing various self-supervised learning, Neural Architecture Search and graph representation learning techniques.     
\end{IEEEbiography}

\vfill

\end{document}


\title{Supplementary material}
\maketitle

\author{Tong Xu,~\IEEEmembership{}
        Micol Spitale,
        Hao Tang,
        Lu Liu，
        Hatice Gune, 
        and
        Siyang Song$^*$

\IEEEcompsocitemizethanks{\IEEEcompsocthanksitem Tong Xu, Lu Liu and Siyang Song are with the School of Computing and Mathematical Sciences, University of Leicester, Leicester, LE2 7RH, United Kingdom.
E-mail:  l.liu@leicester.ac.uk, ss1535@leicester.ac.uk \\
($^*$ Corresponding Author: Siyang Song, E-mail: ss1535@leicester.ac.uk) \\
\indent Micol Spitale, Hatice Gunes and Siyang Song are with the AFAR Lab, Department of Computer Science and Technology, University of Cambridge, Cambridge, CB3 0FT, United Kingdom.
E-mail:  ms2871@cam.ac.uk, Hatice.Gunes@cl.cam.ac.uk, ss2796@cam.ac.uk \\
\indent Hao Tang is with the Department of Information Technology and Electrical Engineering, ETH Zurich,  Zurich 8092, Switzerland. E-mail: hao.tang@vision.ee.ethz.ch}

\thanks{Manuscript received April 19, 2021; revised August 16, 2021.}}

\markboth{Journal of \LaTeX\ Class Files,~Vol.~14, No.~8, August~2021}%
{Shell \MakeLowercase{\textit{et al.}}: A Sample Article Using IEEEtran.cls for IEEE Journals}

\section{Contraction Mapping Theory}
\begin{definition}
\label{definition}
For a non-empty complete metric space $(X, d)$, let the function $\{T: X \rightarrow X\}$ be a map on $X$. Then, the function $T$ is Lipschitz continuous if there exists a constant $k \geq 0$ that meets the equation \ref{eq:Lipschitz}:
\begin{equation}
    d(T(x), T(y)) \leq k \cdot d(x, y)
\label{eq:Lipschitz}
\end{equation}
where $k$ is called the \textbf{Lipschitz constant} of the function $T$, denoted as $k = \mathrm{Lip}(T)$. If $0 \leq k < 1$, the function $T$ is defined as a \textbf{contraction mapping}. 
\end{definition} 

\begin{definition} 
\label{thm:jacobian} 
As presented in \cite{geometric}, let the function $g: \mathbb{R}^n \rightarrow \mathbb{R}^m$ be differentiable and Lipschitz continuous under the $p$-norm metric $\|\cdot\|_p$, where $J_g(\mathbf{x})$ denotes the Jacobian matrix of the function $g$ at the point $\mathbf{x}$. Then, $\mathrm{Lip}_p(g) = \sup_{\mathbf{x} \in \mathbb{R}^n} \|J_g(\mathbf{x})\|_p$ (\textbf{i.e., the Lipschitz constant of a function $g$ equals to the p-norm upper-bound of its Jacobian matrix}), where $\mathrm{Lip}_p(g)$ denotes the $p$-norm based Lipschitz constant; $\sup$ denotes the upper-bound; and $\|J_g(\mathbf{x})\|_p$ is the induced $p$-norm operator on $J_g(\mathbf{x})$.
\end{definition}



\vspace{-3mm}
\section{Reversibility of REGNN}

\noindent As stated in the main manuscript, the reversibility of the proposed REGNN is conditioned on that the forward propagation function $\varphi$ defined in Eqa.~10 of the main manuscript is a contraction mapping. We follow the main manuscript to define $g$ as the combination of the edge updating and the message passing functions for vertex updating during the forward propagation, which is an unbounded function. Consequently, the norm of its Jacobian matrix $J_g(\mathbf{x})$ is also unbounded (please check Eqa.~\ref{eq:20}), i.e., the function $g$ is neither Lipschitz continuous nor a contraction mapping. Therefore, \textbf{the goal is to impose constraints on $g$ to obtain the function $\varphi$ ensuring the propagation of the REGNN to be a contraction mapping, i.e., Lipshitz constant $\^{Lip}_p(\varphi)$ exists and satisfies $\^{Lip}_p(\varphi) < 1$.}

Since the propagation of the REGNN involves computing node and edge features, we first define the following terms for the following derivation:
\begin{align}
\label{eq:terms}
    V^{n-1}_i \in \mathbb{R}^{K \times D} &= \Big[{\Big{\Vert}}_{k \in \mathcal{N}_i} v^{n-1}_k \Big], 
    E^0_i \in \mathbb{R}^{P \times K} &= \Big[{\Big{\Vert}}_{k \in \mathcal{N}_i} e^0_{k, i} \Big]\\
    A^n_i \in \mathbb{R}^{1 \times K} &= \Big[{\Big{\Vert}}_{k \in \mathcal{N}_i} a^n_{k, i} \Big], 
    E^n_i \in \mathbb{R}^{P \times K} &= \Big[{\Big{\Vert}}_{k \in \mathcal{N}_i} e^n_{k, i} \Big]
\end{align}
where $\Big{\Vert}$ is the concatenation operation; please refer the main manuscript for the definition of $v^{n-1}_k$, $e^{0}_{k, i}$, $a^n_{k, i}$ and $e^{n}_{k, i}$. As a result, $V^{n-1}_i$ represents all adjacent vertices of the $v^{n-1}_i$; $E^0_i$ represents all initial edge features connected with $v^{0}_i$; $A^n_i$ represents all coefficients $a^n_{k, i}$ related with $v^{n-1}_i$; and $E^0_i$ represents all updated edge features connected with $v^{n-1}_i$.

According to the Definition~\ref{thm:jacobian}, the Lipschitz constant equals to the upper-bound of the function's $p$-norm (i.e., $\mathrm{Lip}_p(g) = \sup_{v} \|J_{g}(v)\|_p$, where $v$ is the input node feature of the function $g$). In this paper, we set $p$ as $2$ ($2$-norm). \textbf{Then, the goal is reformulated as constrain the function $g$ as the function $\varphi$, allowing $\|J_{\varphi}\|_2$ to have an upper bound $\sup(\|J_{\varphi}\|_2) < 1$}. In this regard, we first compute its Jacobian matrix $J_{g}(v)$. In the main manuscript, the function $g$ is represented as:
\begin{equation}
    g(v_i^{n-1}) = \*{W}^n_e E^{n}_i V^{n-1}_i
\end{equation}
where $ \*{W}^n_e$ is a learnable weight matrix; the propagation of the node feature $v_i^{n-1}$ involves its adjacent nodes, and thus the gradient exists between $g(v^{n-1}_i)$ and the adjacent nodes of $v^{n-1}_i$. Subsequently, we represent the component $J_{g}^{i,j}$ of the Jacobian matrix $J_{g}$ as:
\begin{equation}
\label{eq:total_derivative}
    J_{g}^{i,j}=\frac{\partial g(v^{n-1}_i)}{\partial v^{n-1}_j}=
    (\*{W}^n_e \cdot e^n_{i, j}) I + 
    {V^{n-1}_i}^{\top} \frac{\partial (\*{W}^n_e E^n_i)}{\partial v^{n-1}_j}
\end{equation}
According to the subadditivity of matrix norm: $\|A + B\|_p \leq \|A\|_p + \|B\|_p$, we can obtain $\|J_{g}\|_2 = \big\|\dfrac{\partial g(v^{n-1}_i)}{\partial v^{n-1}_j}\big\|_2 \leq \|(\*W_{e} \cdot e^{n-1}_{i, j}) I\|_2 + \big\|{V^{n-1}_i}^{\top} \dfrac{\partial (\*{W}^n_e E^{n-1}_i)}{\partial v^{n-1}_j}\big\|_2$. Since all the elements in $e^n_{i,j}$ and $\*{W}^n_e$ are less or equaling to 1 while $\^{sum}(\*{W}^n_e)=1$, each element in $\*{W}^n_e \cdot e^{n}_{i,j}$ is also less than or equaling to one. \textbf{As a result, the $2$-norm of the term $(\*{W}^n_e \cdot e^n_{i, j}) I$ has the upper bound $\|I\|_2 = 1$}.

Meanwhile, the term ${V^{n-1}_i}^{\top} \dfrac{\partial (\*{W}^n_e E^n_i)}{\partial v^{n-1}_j}$ can be further represented as:
\begin{equation}
\label{eq:14}
    \frac{\partial (\* W_e E^n_i)}{\partial v^{n-1}_j} = 
    \displaystyle \sum_{d=1}^{D} {\*W^n_e}_d \frac{\partial {E^n_i}_{d:}}{\partial  v^{n-1}_j}
\end{equation}
where ${\*W^n_e}_d$ is the $d$-th element in $\*W^n_e$, and ${E^n_i}_{d:}$ denotes the $d$-th row of the edge feature set $E^n_i$ (i.e., the $d$-th dimension of all features in $E^n_i$). 
As a weighted combination of $D$ partial derivatives, $\dfrac{\partial (\* W_e E^n_i)}{\partial v^{n-1}_j}$ has the upper bound of the $2$-norm $\sup\big(\big\|\dfrac{\partial {E^n}_{d:}}{\partial v^{n-1}_j}\big\|_2\big)$. As a result, the upper-bound of the term ${V^{n-1}_i}^{\top} \dfrac{\partial (\*{W}^n_e E^n_i)}{\partial v^{n-1}_j}$ can be equivalently represented as the upper bound of the ${V^{n-1}_i}^{\top} \dfrac{\partial {E^n_i}_{d:}}{\partial v^{n-1}_j}$ (attributed to their identical upper bounds of the $2$-norm). Subsequently, the term ${V^{n-1}_i}^{\top} \dfrac{\partial {E^n_i}_{d:}}{\partial v^{n-1}_j}$ can be decomposed by the chain rule as:
\begin{equation}
\label{eq:15}
\begin{split}
     &{V^{n-1}_i}^{\top} \frac{\partial {E^n_i}_{d:}}{\partial v^{n-1}_j} = \\
    &\frac{\partial {E^n_i}_{d:}}{\partial (A^n_i \circ {E^0_i}_{d:})}\ \ 
    \frac{\partial (A^n_i \circ {E^0_i}_{d:})}{\partial A^n_i}\ \ 
    \left({V^{n-1}_i}^{\top}\ \frac{\partial A^n_i}{\partial v^{n-1}_j}\right)       
\end{split}
\end{equation}
where $A$ is defined in Eq. \ref{eq:terms} and $\circ$ denotes element-wise product. According to \cite{geometric}, the composition of functions $f\circ h$ has the property $\mathrm{Lip}(f\circ h) \leq \mathrm{Lip}(f) \mathrm{Lip}(h)$, and thus $\^{sup}(J_{f\circ h}) \leq \^{sup}(J_f) \^{sup}(J_h)$. This suggests that the $2$-norm upper bound of the ${V^{n-1}_i}^{\top} \dfrac{\partial {E^{n-1}_i}_{d:}}{\partial v^{n-1}_j}$ (i.e., $\sup\left(\big\|\dfrac{\partial {E^n}_{d:}}{\partial v^{n-1}_j}\big\|_2\right)$) is decided by the three terms defined in Eqa.~\ref{eq:15}.

For \textbf{the first term} $\dfrac{\partial {E^n_i}_{d:}}{\partial (A^n_i \circ {E^0_i}_{d:})}$, it is an normalization operation, and thus we can also represent it as $ \dfrac{\partial \mathrm {norm}(\mathbf z)}{\partial \mathbf z}$. This operation can then has the following derivative form:
\begin{equation}
\begin{split}
    \dfrac{\partial \mathrm {norm}(\mathbf z)}{\partial \mathbf z} &= \dfrac{1}{(\mathbf{1}^\top \mathbf z)^2}(\mathbf {1}^\top \mathbf z \ \mathrm {diag}(\mathbf z) - \mathbf {zz}^\top)\\
    &= \dfrac{\mathrm {diag}(\mathbf{z})}{(\mathbf{1}^\top \mathbf {z})} - \dfrac{\mathbf z}{\mathbf{1}^\top \mathbf {z}} \cdot \dfrac{\mathbf z^\top}{\mathbf{1}^\top \mathbf {z}}    
\end{split}
\end{equation}
where $\* 1$ is a vector where all elements have the value 1, and $\^{diag}()$ is the diagonal matrix.
Let $\mathbf Z=\dfrac{\mathrm{diag}(\mathbf{z})}{(\mathbf{1}^\top \mathbf {z})} - \dfrac{\mathbf z}{\mathbf{1}^\top \mathbf {z}} \cdot \dfrac{\mathbf z^\top}{\mathbf{1}^\top \mathbf {z}}$. 
For any vector $\mathbf a$, $\mathbf{a^\top}\mathbf Z \mathbf a=\mathrm{Var}(\mathbf a) \geq 0$, which means that $\mathbf Z$ is a positive semi-definite matrix and $\lambda(\mathbf Z) = \sigma(\mathbf Z) \geq 0$, where $\lambda()$ denotes eigenvalues and $\sigma()$ denotes singular values.
Furthermore, $\mathbf {tr}\left(\dfrac{\mathrm{diag}(\mathbf z)}{\mathbf{1}^\top \mathbf z}\right) = 1$, 
$0 < \mathbf {tr}\left(\dfrac{\mathbf z}{\mathbf{1}^\top \mathbf {z}} \cdot \dfrac{\mathbf z^\top}{\mathbf{1}^\top \mathbf {z}}\right) \leq 1$, 
thus $\mathbf {tr}(\mathbf Z) \leq 1$. 
Since $\lambda (\* Z) \geq 0$ and $\* Z$ is a symmetric matrix, $\sigma (\* Z) = \lambda (\* Z)$. $\sigma(\* Z) < \*{tr(Z)} \leq 1$, then $\max (\sigma(\mathbf Z)) = \max(\lambda(\mathbf Z)) \leq 1$. Here, $\mathbf{tr}$ denotes the trace. As the $2$-norm is defined as the maximum singular value of a matrix, \textbf{we come to the conclusion that the upper-bound the the first term $\^{sup}\left(\left\|\dfrac{\partial {E^n_i}_{d:}}{\partial (A^n_i \circ {E^0_i}_{d:})}\right\|_2\right) \leq 1$. }

According to the derivative rule of the matrix, \textbf{the second term} $\dfrac{\partial (A^n_i \circ {E^0_i}_{d:})}{\partial A^n_i}$ can be represented as:
\begin{equation}
    \frac{\partial (A^n_i \circ {E^0_i}_{d:})}
    {\partial A^n_i}=\mathrm{diag}({E^0_i}_{d:})
\end{equation}
As the edge feature set $E^0_i$ is normalized, where $\sum{E^0_i}_{d:}=1$, we can obtain that $\max\left(\sigma\left(\mathrm{diag}({E^0_i}_{d:})\right)\right) = \max({E^0_i}_{d:}) \leq 1$. \textbf{Consequently, the upper-bound of the second term can be obtained as $\^{sup}\left(\left\|\dfrac{\partial (A^n_i \circ {E^0_i}_{d:})}{\partial A^n_i}\right\|_2\right)=1$.}

Finally, \textbf{the third term} $\left({V^{n-1}_i}^{\top}\ \dfrac{\partial A^n_i}{\partial v^{n-1}_j}\right)$ can be computed as:
\begin{equation}
\label{eq:third}
    {V^{n-1}_i}^{\top}\dfrac{\partial A^n_i}{\partial v^{n-1}_j}={V^{n-1}_i}^{\top}A^{(i)}\left(B_{j,i}{V^{n-1}_i} \*U^\top + {V^{n-1}_i} \*U\delta_{i,j}\right)
\end{equation}
where $A^{(i)}=\mathrm {diag}(A^n_{i})-{A^n_{i}}^\top A^n_{i}$; $B_{j,i}$ is an matrix where the component located at $i_\text{th}$ row and $j_\text{th}$ column is 1, while the rest components are set as 0; $\* U = \*W^n_q {\*W^n_m}^{\top}$, and $\delta_{i,j}$ is the Kronecker delta. Part of this derivation procedure is obtained according to \cite{lipattn}.
Here, we discuss the third term in two cases:
\begin{itemize}
\item \textbf{Case 1: $i = j$}. The third term $\left({V^{n-1}_i}^{\top}\ \dfrac{\partial A^n_i}{\partial v^{n-1}_j}\right)$ can be refined as:
\begin{equation}
\label{eq:20}
    a^n_{i,i}\Big(v^{n-1}_i - \displaystyle\sum_k a^n_{ik} v^{n-1}_k\Big) {v^{n-1}_i}^\top \*U^\top+ 
    {V^{n-1}_i}^\top A^{(i)}V^{n-1}_i \* U
\end{equation}
The value of the term ${V^{n-1}_i}^\top A^{(i)}V^{n-1}_i \* U$ is unlimited as its 2-norm is conditioned on the value of the input vertex feature set $V^{n-1}_i$ which has no upper-bound. Subsequently, there is no Lipschitz constant for the original function $g$ as the 2-norm of its Jacobian matrix $\|J_g(\mathbf{x})\|_2$ is not upper-bounded, and therefore $g$ is not a contraction mapping. To address this problem, we modify the function $g$ to bound the input for each REGNN layer, which employs an sigmoid activation function to process the input vertex features $V$ in each REGNN layer, and thus the input is limited within an interval $[0,1]$. Such bounded inputs would also ensure the term ${V^{n-1}_i}^\top A^{(i)}V^{n-1}_i \* U$ to have an upper-bound. According to Definition \ref{definition}, the function $\varphi$ would be Lipschitz continuous. Given that the input space is bounded, we are able to obtain an upper-bound on the term ${V^{n-1}_i}^\top A^{(i)}V^{n-1}_i \* U$ as $\|\* U\|_2$. Similarly, since $A^n_{i,j} < 1$, \textbf{the upper bound of the term $ a^n_{i,i}\Big(v^{n-1}_i - \displaystyle\sum_k a^n_{ik} v^{n-1}_k\Big) {v^{n-1}_i}^\top \*U^\top$ is $\|\* U\|_2$.}

\item \textbf{Case 2: $i \neq j$}. In this case, only one term exists, which is defined as:
\begin{equation}
    a^n_{i,j}\Big(v^{n-1}_j - \displaystyle\sum_k a^n_{i,k} v^{n-1}_k\Big) {v^{n-1}_i}^\top \*U^\top
\end{equation}
For the same reason as (1), the upper bound is $\|\* U\|_2$. Consequently, \textbf{we conclude that $\left\|{V^{n-1}_i}^{\top}\dfrac{\partial A^n_i}{\partial v^{n-1}_j}\right\|_2 < 2\|\* U\|_2$.} 
\end{itemize}

In summary, the proposed input bounding strategy lead to the upper-bounds $1$, $1$, and $2\|\* U\|_2$ for the three terms decomposed in Eqa.~\ref{eq:15}. According to the subadditivity of the norm, the upper-bound of the second term in Eq. \ref{eq:total_derivative} is $2\|\* U\|_2$, while the upper-bound of the first term in Eq. \ref{eq:total_derivative} is $\|I\|_2$. Consequently, we ultimately obtain the 2-norm upper-bound of the Jacobian matrix for the function $g$ as $\|I\|_2 + 2\|\* U\|_2$ (i.e., $\sup \|J_g(\mathbf{v})\|_2 < \|I\|_2 + 2\|\* U\|_2)$. To make the $g$ to be a contraction mapping (i.e., $\sup \|J_g(\mathbf{v})\|_2 < 1$), we modify it as the function $\varphi$ to obtain a contraction mapping $\varphi$ as: $\varphi(v_i^{n-1}) = \dfrac{g(\text{Sig}(v_i^{n-1}))}{{\|I\|_2 + 2\|\* U\|_2}}$. Subsequently, the 2-norm upper-bound of its Jacobian matrix is less than 1 (i.e., $\sup \|J_{\varphi}(\mathbf{v})\|_2 < 1$ ), which means thus Lipschitz constant $\mathrm{Lip}_2(\varphi) < 1$. Based on the definition \ref{definition} and definition \ref{thm:jacobian}, defining the forward propagation as the function $\varphi$ allows the REGNN layer to be reversible.

\section{Details of the reproduced baselines}

\textbf{Huang et al. (S) \cite{CGAN-BMVC}:} This method employs a two-stage conditional GAN. In the first stage, feature vectors composed of $17$ different facial action units of the speaker's facial behaviour are used as the condition to predict the listener's facial landmarks. Then, in the second stage, the obtained facial landmarks are employed as the condition to generate the corresponding facial reaction sequence. Here,  the speaker facial behaviour from time $t-9$ to $t$ is used to predict the $t_\text{th}$ facial reaction frame. However, this method is unable to predict the first ten facial reaction frames. Therefore, for our experiments, the input size is set to $(750, 25)$, while the output size is set to $(740, 25)$.

\textbf{Huang et al. (C) \cite{CGAN-CVPR}:}  Building upon \cite{CGAN-BMVC}, this method utilizes a feature vector composed of $8$ distinct facial expressions of the speaker behaviour as a conditioning input. The final facial reaction is generated using a two-stage GAN similar to \cite{CGAN-CVPR}. As both \cite{CGAN-BMVC} and \cite{CGAN-CVPR} require the inputs from time $t-9$ to $t$ to predict the outputs at time $t$, they cannot make predictions for the first ten frames of the listener's facial reaction. Therefore, in our experiments, their input size is set to $(750, 25)$, and the output is set to $(740, 25)$.

\textbf{UNet \cite{unet}:} UNet is a widely-used model in generative tasks, consisting of the symmetric encoder and decoder components. In this experiment, both the encoder and decoder are set to four layers. The input and output of the network are both time series data with a shape of $(750, 25)$.

\textbf{Ng et al. \cite{ng2022learning}:} It is a VQ-VAE model that predicts a discrete encoding from the input speaker behaviour, representing the corresponding listener's facial reaction. At the inference stage, a transformer-based predictor module is utilized to perform a lookup on the discrete encoding to predict the listener's facial reaction. In our experiments, we customized the size of hidden layers in the model to accommodate the input sequence size of $(750, 25)$ and output size of $(750, 25)$.

\ifCLASSOPTIONcaptionsoff
  \newpage
\fi

{\small
	\bibliographystyle{IEEEtran}
	\bibliography{egbib}
}

\vfill